\documentclass{article}

\PassOptionsToPackage{numbers,sort&compress}{natbib}

\usepackage[preprint]{neurips_2026}

\usepackage[utf8]{inputenc}
\usepackage[T1]{fontenc}
\usepackage[breaklinks=true]{hyperref}
\usepackage{bookmark}
\usepackage{url}
\usepackage{booktabs}
\usepackage{amsfonts}
\usepackage{amsmath}
\usepackage{nicefrac}
\usepackage{microtype}
\usepackage[table]{xcolor}
\usepackage{graphicx}
\usepackage{wrapfig}
\usepackage{tikz}
\usepackage{makecell}
\usepackage{placeins}

\definecolor{VPGGain}{HTML}{34A853}
\newcommand{\gain}[1]{\textcolor{VPGGain}{\textbf{#1}}}

\usepackage{soul}

\usepackage{xspace}

\makeatletter
\DeclareRobustCommand\onedot{\futurelet\@let@token\@onedot}
\def\@onedot{\ifx\@let@token.\else.\null\fi\xspace}

\def\ie{\emph{i.e}\onedot} 
 
\makeatother

\let\origcitep\citep
\renewcommand{\citep}[2][]{\mbox{\origcitep[#1]{#2}}}
\renewcommand{\citet}[2][]{\mbox{\origcitep[#1]{#2}}}

\title{VPG: Visual Prefix Guidance for Autoregressive Image and Video Generation}

\author{%
Xinyao Liao$^{1,2}$ \quad Qiyuan He$^{1}$\thanks{Project lead.} \quad Yicong Li$^{1}$ \quad Jiayin Zhu$^{1}$ \quad \\ \textbf{Xiaoye Qu$^{2}$ \quad Wei Wei$^{2}$ \quad Angela Yao$^{1}$} \\
$^1$National University of Singapore  \quad $^2$Huazhong University of Science \& Technology \\
}

\begin{document}

\maketitle

\begin{figure}[h]
  \centering
  \includegraphics[width=0.9\linewidth]{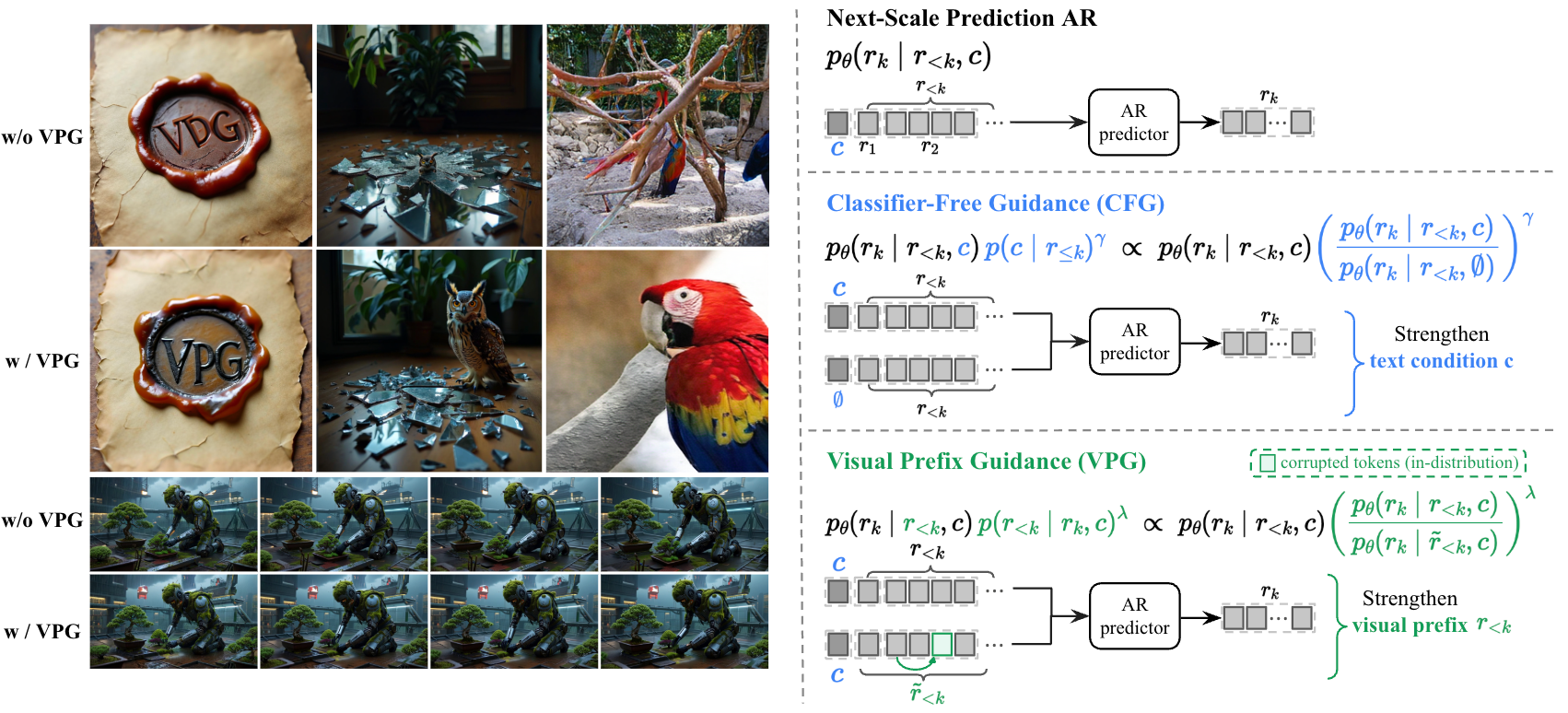}
  \caption{\textbf{Visual Prefix Guidance (VPG) sharpens dependence on the generated visual prefix}, complementing CFG along an axis a frozen model cannot reach. Left: same-prompt, same-seed comparisons w/o vs.\ w/ VPG; the first two columns use Infinity~\citep{infinity} prompts for a ``VPG'' wax seal and an owl among shattered mirrors, the third uses VAR-$d30$~\citep{var} for ``macaw'', and the last two rows use InfinityStar~\citep{InfinityStar} for a moss-covered mech tending rooftop bonsai. Right: CFG sharpens $p(c\mid r_{\le k})^\gamma$, while VPG sharpens $p(r_{<k}\mid r_k,c)^\lambda$. See Appendix~\ref{app:teaser-details}.}
  \label{fig:teaser}
\end{figure}

\begin{abstract}
Autoregressive image and video generators are trained with teacher-forced histories but must sample from their own generated prefixes at inference time, making them vulnerable to exposure bias and prefix drift. Existing remedies either modify training or apply sampling-time guidance aimed primarily at external semantic conditions, such as class labels or text prompts, rather than testing whether a next-step prediction provides strong posterior support for the generated prefix itself. We propose \textbf{Visual Prefix Guidance (VPG)}, a training-free inference-time guidance method for autoregressive image and video generation. VPG improves next-step prediction by contrasting the model's output under the generated prefix with its output under a corrupted prefix, then extrapolating logits toward candidates that strengthen the posterior support of the generated prefix. Across class-conditional image generation with VAR, text-to-image generation with Infinity, and text-to-video generation with InfinityStar, VPG improves generation quality without retraining the base model, reducing FID on VAR by 0.36 on average and improving benchmark performance on both image and video generation.

\end{abstract}

\section{Introduction}
\label{sec:introduction}

Visual autoregressive models~\citep{llamagen, var, infinity, InfinityStar} are becoming a strong alternative to diffusion models~\citep{flux-2-2025,kong2024hunyuanvideo,wan2025wan,liao2025step} for both image and video generation.
Instead of denoising continuous latents through an iterative reverse process, these models generate visual tokens, scales, or latent variables sequentially.
This formulation brings visual synthesis closer to GPT-style sequence modeling~\citep{gpt,gemini}, which 
scale well in language and multimodal
systems~\citep{team2024chameleon,deng2025bagel}.

Autoregressive models recursively generate each new prediction based on the generated history.
We use the term \emph{prefix} to denote this history for general model, independent of the prediction unit, which may be a single token, a token map, or a full image frame.
During training, these models are usually trained with teacher forcing: each next-step prediction is conditioned on the ground-truth prefix, so the model only observes correct histories during optimization. However, during inference, ground-truth prefixes are unavailable, and the model must condition on its own sampled history. The training-inference mismatch creates exposure bias~\citep{scheduled_sampling,wang2020exposure}: early sampling errors
can lead the model into unseen or low-probability prefix states and 
accumulate errors over subsequent predictions.
At inference time, errors in this generated prefix can accumulate and shift later predictions away from the training distribution. In visual generation, such prefix drift often appears as structural errors, object inconsistency, or unstable high-frequency details~\citep{shin2026ssg,huang2024vbench}.

Recent work addresses it from two angles. Training-time methods incorporate generated or perturbed histories during optimization~\citep{rear,scheduled_sampling,xar,professor_forcing,vapi} to make the model more robust against the drifted prefix.
\emph{Sampling-time} guidance modifies the inference rule to mitigate exposure bias by strengthening the textual conditioning component available in the prefix, as in CFG~\citep{classifier_free_guidance} and recent variants based on degraded or self-perturbed references~\citep{karras2024autoguidance,ahn2024pag,hong2024seg,softcfg,shin2026ssg}. Yet both lines of work leave a key question unaddressed: when conditioning on a self-generated visual prefix, how should the next prediction be adjusted to reduce subsequent drift? Exposure bias matters not only because the prefix may contain errors, but because later predictions may become increasingly unsupported by that generated history.

This motivates \textbf{Visual Prefix Guidance (VPG)}, a training-free, drop-in sampling rule for next-scale prediction. In standard visual AR, sampling proceeds from the conditional likelihood $p_\theta(r_k\mid r_{<k}, c)$. VPG adds a different lens: at each step, it favors candidates that increase the posterior support of the generated prefix, $p(r_{<k}\mid r_k, c)$, thereby suppressing unsupported drift before it compounds. 

To realize this prefix-posterior objective, VPG forms a paired prediction contrast between the genuine generated prefix and an
inference-time corrupted prefix (as in Fig.~\ref{fig:corrupted-prefix-surrogate}). The corrupted branch approximates a prefix-marginalized likelihood that a frozen visual AR model does not provide natively, playing the prefix-axis analogue of the null condition in CFG. We instantiate this reference by \emph{same-scale full-embedding replacement}: at each prefix scale, we replace a random fraction of token-position embeddings with full embeddings copied from other positions at the same scale. This yields a weaker prefix branch while preserving the model’s scale-conditional input
statistics. This prefix-posterior view is complementary to condition-based guidance such as CFG
while targeting a different failure mode: unsupported drift from the generated history itself.

We evaluate VPG in 
class-conditional image generation with VAR~\citep{var}, text-to-image generation with Infinity~\citep{infinity}, and text-to-video generation with InfinityStar~\citep{InfinityStar}. Across these settings, 
VPG improves 
prediction quality, 
all without retraining the base generator. On class-conditional VAR, VPG reduces FID by \textbf{0.36} on average across model sizes, and by as much as \textbf{0.63} for VAR-d16. 
On text-to-image generation, 
VPG improves both GenEval Overall and DPG-Bench Overall. On video generation, 
VPG improves the VBench~\citep{huang2024vbench} Overall score by \textbf{0.49}, with gains across all sub-scores.  Our contributions are summarized as follows:
\begin{itemize}
    \item We identify a previously underexplored inference-time objective for visual autoregressive generation %
    by strengthening posterior support for the generated prefix. This 
    improves next-step prediction by directly targeting the 
    accumulation of exposure bias in the prefix.  
    \item We propose Visual Prefix Guidance (VPG), a training-free, drop-in sampling rule for next-scale prediction that realizes this objective with a corrupted-prefix contrast and same-scale full-embedding replacement, requiring no auxiliary head or retraining.
    \item We show that VPG consistently improves autoregressive image and video generation across VAR, Infinity, and InfinityStar, reducing FID on VAR by 0.36 on average and improving benchmark scores on both text-to-image and text-to-video generation.
\end{itemize}


\section{Related work}
\label{sec:related}

\noindent\textbf{Visual autoregressive generation.}
Autoregressive (AR) models are now a competitive paradigm for visual generation~\citep{darn,imagetransformer,igpt,llamagen}. They have been applied to both image and video synthesis. Direct pixel-level autoregression is prohibitively expensive, so modern AR systems instead operate on compressed visual latents, often with discrete tokenizers such as VQ-VAE~\citep{vqvae} and VQ-GAN~\citep{vqgan}. Recent work improves this pipeline through better tokenization~\citep{magvit-v2,fsq,unitok,mar,titok, ye2025infotok}. Other work changes the autoregressive structure itself, for example through coarse-to-fine scale prediction in VAR~\citep{var} and bitwise tokenization with scalable generation in Infinity~\citep{infinity,InfinityStar}.

\noindent\textbf{Exposure bias and prefix mismatch.}
Autoregressive models are commonly trained with teacher forcing, where each prediction conditions on a ground-truth prefix. At inference time, the model must condition on its own sampled prefix, creating exposure bias~\citep{scheduled_sampling,wang2020exposure}. This train--test mismatch has long been studied in sequence modeling. Classical remedies either mix model-generated prefix into training~\citep{scheduled_sampling,pmlr-v15-ross11a,zhang-etal-2019-bridging}, align teacher-forced and free-running dynamics~\citep{professor_forcing}, or train on noisy or revised autoregressive contexts~\citep{rear,xar}. A complementary line of work studies how severe exposure bias actually is, framing it through generalization or explicit error-accumulation analyses~\citep{schmidt-2019-generalization,arora-etal-2022-exposure,wang2020exposure}. In visual generation, recent methods likewise regularize self-generated rollouts or add post-training refinement objectives~\citep{vapi,huang2025selfforcing,liu2026interpd}. These approaches mainly modify training or add post-training optimization, whereas our goal is to improve robustness purely at inference time.

\noindent\textbf{Sampling-time guidance and self-contrasting strategy.}
A complementary line of work studies training-free \emph{sampling-time guidance}. The canonical example is classifier-free guidance (CFG), which combines conditional and unconditional predictions to bias the generation toward an external semantic condition such as a class label or text prompt~\citep{classifier_free_guidance}. Follow-up works replace the unconditional reference with a degraded internal reference. Variants of the reference have been explored from different perspectives, including weaker denoisers for autoguidance~\citep{karras2024autoguidance}, attention manipulation~\citep{ahn2024pag,hong2024seg,he2024aid,he2025conceptrol}, and feature swaps in diffusion models~\citep{zhang2026swap,zhu2026relaxflow}. The analogues in autoregressive visual generation use reweighted conditional predictions or scale-specific self-guidance~\citep{softcfg,shin2026ssg}. These methods show that internal contrast can be a powerful guidance signal, though they are still primarily organized around external condition. In contrast, our work focus on if next-step prediction is supported by the generated prefix itself, giving a direct sampling-time handle on prefix drift. 

\section{Preliminaries}
\label{sec:preliminaries}

\subsection{Next-Scale Visual Autoregressive Models}
\label{sec:prelim-var}

Following VAR~\citep{var} and its bitwise extension Infinity~\citep{infinity}, we model visual generation as a next-scale prediction process. A multi-scale tokenizer maps an image $I$ into $K$ coarse-to-fine residual token maps $R=(r_1,\ldots,r_K)$, where each \emph{scale} corresponds to one stage of the residual decomposition. Thus, $K$ is the total number of residual prediction stages, ordered from the coarsest map $r_1$ to the finest map $r_K$.
Each residual token map is written as $r_k\in\mathcal{V}^{h_k\times w_k}$, where $\mathcal{V}$ denotes the token space at each spatial location. In VAR, entries of $r_k$ are VQ-VAE~\citep{vqvae} codebook indices; in Infinity, they are represented by BSQ codes~\citep{zhao2024bsq}.
The AR model factorizes generation over these residual scales:

\begin{equation}
  p_\theta(R\mid c) = \prod_{k=1}^{K} p_\theta(r_k \mid r_{<k}, c),
  \label{eq:var-factorization}
\end{equation}
where $c$ is an external condition, such as a class label or text prompt.  The prefix $r_{<k}$ contains all previous residual scales. During training, it comes from ground-truth token maps. During inference, it is replaced by the model's 
sampled prefix, which for 
simplicity, we also 
denote 
as $r_{<k}$.

Given condition $c$ and a generated prefix $r_{<k}$, the transformer outputs the scale-$k$ logit tensor $\ell_k(c,r_{<k})$. These logits define one token distribution at each spatial location of the next $h_k\times w_k$ residual map. Sampling them produces $\hat r_k\in\mathcal{V}^{h_k\times w_k}$.  The sampled token map is then de-quantized and added to the partial latent reconstruction:
\begin{equation}
  z_k = Q_k^{-1}(\hat r_k), \qquad
  \hat f_k = \hat f_{k-1} + U_k(z_k), \qquad
  \hat f_0 = 0,
  \label{eq:latent-accumulation}
\end{equation}
where $Q_k^{-1}$ is the scale-specific de-quantizer and $U_k$ upsamples the residual feature map to the latent resolution. After the final scale, the decoder maps the accumulated latent to the image, $\hat I = D(\hat f_K)$.

\subsection{Classifier-Free Guidance in Visual Autoregression}


\label{sec:prelim-ar-cfg}

The factorization in Eq.~\eqref{eq:var-factorization} exposes two conditioning inputs: the external condition $c$ and the autoregressive prefix $r_{<k}$. We use this distinction to organize sampling-time guidance methods by which input is held fixed and which input is contrasted to form a guidance direction (Tab.~\ref{tab:guidance-axes}).

\begin{table}[t]
  \centering
  \caption{\textbf{Guidance axes in conditional visual autoregression.} In $p_\theta(r_k\mid r_{<k},c)$, the \emph{fixed axis} is unchanged between the two compared branches being, while the \emph{contrasted axis} is the conditioning variable that is perturbed to form the guidance signal. CFG guides the external condition $c$, while VPG guides the generated prefix $r_{<k}$. The two contrasts can be composed to guide both axes.}
  \label{tab:guidance-axes}
  \begin{tabular}{llll}
    \toprule
    Method & {Fixed axis} & Contrasted axis & Guided dependence \\
    \midrule
    CFG & $r_{<k}$ & $c$ vs.\ $\emptyset$ & external condition $c$ \\
    VPG & $c$ & $r_{<k}$ vs.\ $\tilde r_{<k}$ & generated prefix $r_{<k}$ \\
    CFG+VPG & {--} & both contrasts & $c$ and $r_{<k}$ \\
    \bottomrule
  \end{tabular}
\end{table}

Sampling directly from the next-scale conditional $p_\theta(r_k\mid r_{<k}, c)$ in Eq.~\eqref{eq:var-factorization} 
does not necessarily lead to generations 
faithful to the external condition $c$; the sampler may favor $r_k$ with high prefix-prior likelihood $p_\theta(r_k\mid r_{<k})$ but only weakly support $c$. Classifier-free guidance (CFG)~\citep{classifier_free_guidance} addresses this at sampling time by sharpening the \emph{posterior compatibility} of the external condition, $p(c\mid r_{\le k})$, with the prefix sequence after appending $r_k$, where $r_{\le k}=(r_{<k},r_k)$. This is {realized by} exponentiating the compatibility term in the augmented (unnormalized) conditional
\begin{equation}
  p_\theta^{\mathrm{CFG}}(r_k\mid r_{<k}, c)
  \;\propto\;
  p_\theta(r_k\mid r_{<k}, c)\,p(c\mid r_{\le k})^{\gamma},
  \label{eq:cfg-augmented}
\end{equation}
where $\gamma\ge 0$ is the guidance strength. Eq.~\eqref{eq:cfg-augmented} is written as a classifier-guidance objective because it seems to require the posterior $p(c\mid r_{\le k})$. However, applying Bayes' rule along the AR chain removes the need for an explicit classifier:
\[
  p(c\mid r_{\le k})
  \;\propto_{r_k}\;
  \frac{p_\theta(r_k\mid r_{<k},c)}
       {p(r_k\mid r_{<k})}.
\]
Therefore CFG samples from the same augmented distribution by reweighting the next-scale conditional with a conditional/unconditional likelihood ratio,
\[
  p_\theta^{\mathrm{CFG}}(r_k\mid r_{<k}, c)
  \;\propto\;
  p_\theta(r_k\mid r_{<k}, c)
  \left(
    \frac{p_\theta(r_k\mid r_{<k},c)}
         {p(r_k\mid r_{<k})}
  \right)^\gamma,
\]
up to normalization over $r_k$ (full derivation in App.~\ref{app:cfg-derivation}). The remaining inaccessible term is the unconditional predictive $p(r_k\mid r_{<k})$. Classifier-free training supplies a learned null condition $\emptyset$ to approximate this term, \ie $p_\theta(r_k\mid r_{<k})\approx p_\theta(r_k\mid r_{<k},\emptyset)$. Sampling from Eq.~\eqref{eq:cfg-augmented} then translates directly to the logit-space extrapolation rule.
\begin{equation}
  \ell^{\mathrm{CFG}}_{k}
  \;=\;
  \ell_k^c
  + \gamma\bigl(\ell_k^c-\ell_k^\emptyset\bigr)
  \;=\;
  (1+\gamma)\,\ell_k^c
  - \gamma\,\ell_k^\emptyset.
  \label{eq:cfg-logits}
\end{equation}
Above, $\ell_k^c =\ell_k(c,r_{<k})$ and $\ell_k^\emptyset = \ell_k(\emptyset,r_{<k})$ denote the scale-$k$ logits under the condition and the null-condition branches with the same generated prefix, respectively, and $\gamma$ is the CFG scale. Increasing $\gamma$ pushes the sampler toward next-scale candidates that maximize $p(c\mid r_{\le k})$.

In the autoregressive conditional $p_\theta(r_k\mid r_{<k},c)$, the standard CFG only guides the external condition $c$, by contrasting $c$ with $\emptyset$ while keeping the generated prefix fixed.
This view casts CFG as a log-ratio guidance over one conditioning axis.

\section{Method}
\label{sec:method}

\begin{figure}[t]
  \centering
  \includegraphics[width=\linewidth]{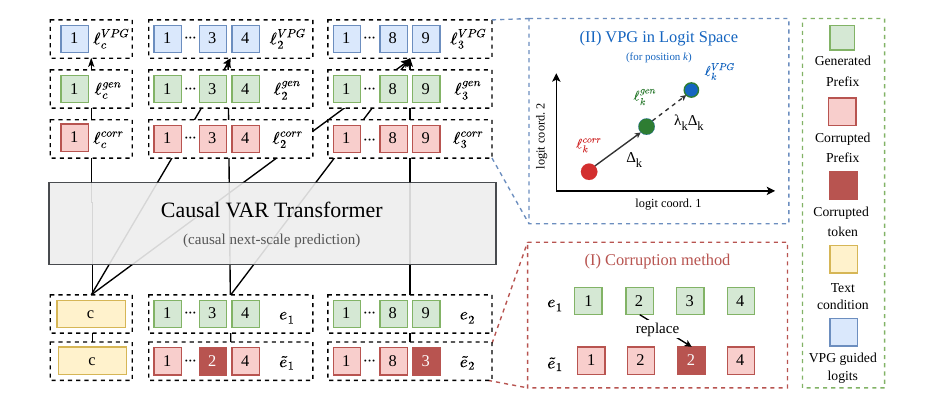}
  \caption{\textbf{Visual Prefix Guidance (VPG) framework.}
  VPG contrasts logits from the generated prefix $r_{<k}$ and same-scale corrupted prefix $\tilde r_{<k}$ using the same frozen transformer and condition $c$.
  The resulting direction is extrapolated before sampling the next token map $\hat r_k$.}
  \label{fig:vpg-system}
\end{figure}

\subsection{Visual Prefix Guidance as Log-Ratio Guidance}
\label{sec:method-vpg}

Visual Prefix Guidance (VPG) starts from the inference-time setting in which the next-scale model must sample from $p_\theta(r_k\mid r_{<k}, c)$, where $r_{<k}$ is the generated prefix available at step $k$. Under exposure bias, later predictions can drift because they are weakly supported by this generated history. VPG therefore changes the sampling objective: among plausible next-scale candidates, it favors those that provide a stronger posterior support for the prefix $r_{<k}$,~\ie~
$p(r_{<k}\mid r_k, c)$.

This offers a training-free way to target prefix drift. Instead of modifying the model or revising the prefix, VPG augments the next-step conditional with a prefix-compatibility term:
\begin{equation}
  p_\theta^{\mathrm{VPG}}(r_k\mid r_{<k}, c)
  \;\propto\;
  p_\theta(r_k\mid r_{<k}, c)\,
  p(r_{<k}\mid r_k, c)^{\lambda},
  \label{eq:vpg-distribution}
\end{equation}
where $p_\theta^{\mathrm{VPG}}$ denotes the VPG-guided sampling distribution and $\lambda\ge 0$ is the prefix-guidance strength. This mirrors the compatibility-augmentation view of CFG (Sec.~\ref{sec:prelim-ar-cfg}), but acts on a different conditioning axis: CFG sharpens agreement with the external condition $c$, while VPG sharpens agreement with the generated visual prefix (Tab.~\ref{tab:guidance-axes}). Fig.~\ref{fig:vpg-system} illustrates the resulting two-pass pipeline and the logit-space extrapolation used before sampling the next scale.

Eq.~\eqref{eq:vpg-distribution} is written as a prefix-classifier objective because it seems to require the posterior $p(r_{<k}\mid r_k,c)$. However, applying Bayes' rule along the prefix axis removes the need for an explicit classifier:
\begin{equation}
  \begin{aligned}
    \!\!\! p(r_{<k}\mid r_k, c)
    &\;\propto_{r_k}\;
    \frac{p_\theta(r_k\mid r_{<k}, c)}
         {p(r_k\mid c)}, \;\;
    \text{where}\;\; p(r_k\mid c)
    = \int\!\! p_\theta(r_k\mid r_{<k}, c)\, p(r_{<k}\mid c)\,\mathrm{d}\mu(r_{<k}).
  \end{aligned}
  \label{eq:vpg-bayes}
\end{equation}
Here $\mu$ denotes the base measure over prefixes; for discrete token prefixes, the integral reduces to a sum under the counting measure.
Therefore ,VPG samples from the same augmented distribution by reweighting the next-scale conditional with a prefix-conditional/prefix-marginalized likelihood ratio,
\[
  p_\theta^{\mathrm{VPG}}(r_k\mid r_{<k}, c)
  \;\propto\;
  p_\theta(r_k\mid r_{<k}, c)
  \left(
    \frac{p_\theta(r_k\mid r_{<k},c)}
         {p(r_k\mid c)}
  \right)^\lambda,
\]
up to a normalization term over $r_k$ (see full derivation in App.~\ref{app:vpg-derivation}). 

The remaining intractable term is the prefix-marginalized likelihood $p(r_k\mid c)$, which is exactly because it requires marginalizing over all possible prefixes. 
Unlike CFG, a frozen visual AR model is not trained with a dedicated null-prefix input. 
Replacing the prefix with random tokens may mimic a null condition but can also introduce out-of-distribution inputs. 
We therefore use a tractable surrogate whose prefix remains compatible with the pretrained model while carrying less prefix-specific information than the clean prefix $r_{<k}$. 
Specifically, VPG constructs a weak-prefix reference by evaluating the frozen model under an inference-time corrupted prefix:
\begin{equation}
    p_{\mathrm{ref}}(r_k\mid c)
    :=
    p_\theta(r_k\mid \tilde r_{<k},c).
\end{equation}
The corrupted prefix plays a role analogous to the null condition in CFG, but along the autoregressive prefix axis rather than the external-condition axis. 
Although such a surrogate may be inconsistent, as can also occur for null-condition guidance under model misspecification~\citep{grunwald2007suboptimal, classifier_free_guidance}, we empirically find that it help improve conditioning ability on the prefix. 
We provide further discussion in App.~\ref{app:vpg-derivation}.

Sampling from Eq.~\eqref{eq:vpg-distribution} then translates directly to a logit-space extrapolation rule. Let $\ell_k^{\mathrm{gen}}\equiv\ell_k(c,r_{<k})$ and $\ell_k^{\mathrm{corr}}\equiv\ell_k(c,\tilde r_{<k})$ denote the scale-$k$ logits from the generated-prefix and corrupted-prefix branches respectively. The prefix-guided logit $ \ell^{\mathrm{VPG}}_{k}$ is then given as
\begin{equation}
  \ell^{\mathrm{VPG}}_{k}
  \;=\;
  \ell_k^{\mathrm{gen}}
  + \lambda\bigl(
    \ell_k^{\mathrm{gen}}
    -
    \ell_k^{\mathrm{corr}}
  \bigr)
  \;=\;
  (1+\lambda)\,\ell_k^{\mathrm{gen}}
  - \lambda\,\ell_k^{\mathrm{corr}},
  \label{eq:vpg-logits}
\end{equation}
Eq.~\eqref{eq:vpg-logits} is applied position-wise before sampling the next-scale tokens: for VAR-style models, over the parallel prediction heads at all $h_k w_k$ sites; for Infinity-style models, over the corresponding bitwise logits. Note that VPG changes only the inference-time sampling distribution and introduces no auxiliary model, learned guidance head, or retraining objective.

\paragraph{Composition with CFG.}
CFG and VPG are composed directly in logit space. We first apply CFG separately to the genuine-prefix and corrupted-prefix branches,
\begin{align}
  g_k^{\mathrm{gen}}
  &=
  \ell_k(c,r_{<k})
  + \underbrace{\gamma\bigl(\ell_k(c,r_{<k}) - \ell_k(\emptyset,r_{<k})\bigr)}_{\text{CFG on genuine prefix}}, \\
  g_k^{\mathrm{corr}}
  &=
  \ell_k(c,\tilde r_{<k})
  + \underbrace{\gamma\bigl(\ell_k(c,\tilde r_{<k}) - \ell_k(\emptyset,\tilde r_{<k})\bigr)}_{\text{CFG on corrupted prefix}},
\end{align}
where $\gamma$ is the CFG scale, and $g_k^{\mathrm{gen}}$ and $g_k^{\mathrm{corr}}$ denote the CFG-guided intermediate logits from the generated-prefix and corrupted-prefix branches. We then apply VPG between these two branch logits:
\begin{equation}
  \ell^{\mathrm{CFG+VPG}}_{k}
  \;=\;
  \underbrace{g_k^{\mathrm{gen}}}_{\text{CFG-guided base}}
  + \underbrace{\lambda\bigl(
    g_k^{\mathrm{gen}}
    -
    g_k^{\mathrm{corr}}
  \bigr)}_{\text{VPG}}.
  \label{eq:cfg-vpg-logits}
\end{equation}
Eq.~\eqref{eq:cfg-vpg-logits} first strengthens agreement with the external semantic condition $c$ on each prefix branch, then strengthens agreement with the genuine generated prefix. The corresponding likelihood interpretation and derivation are provided in App.~\ref{app:cfg-vpg-composition}.

\subsection{Constructing a Corrupted Prefix}
\label{sec:method-prefix-substitution}

The corrupted prefix $\tilde r_{<k}$ enters VPG as a single-sample surrogate for the prefix-marginalized predictive $p(r_k\mid c) = \mathbb{E}_{r_{<k}\sim p(\cdot\mid c)}[p_\theta(r_k\mid r_{<k}, c)]$ in Eq.~\eqref{eq:vpg-bayes}. The ideal surrogate is a draw from $p(r_{<k}\mid c)$, that is content-independent of the generated prefix, while ensuring every other model input (scale geometry, position encoding, $c$, weights) stays within the training distribution.

We approximate such a draw at zero training cost by \emph{same-scale full embedding replacement}: across the prefix, a uniformly random fraction $n_p$ of sites have their full embedding (visual code with its scale-position encoding) replaced with that of another site at the same scale, where $n_p$ is the \emph{corruption fraction} (See ablations in Sec.~\ref{sec:experiments-var-sensitivity}). The replacement embeddings come from the model itself, so scale-conditional statistics are preserved, while the content-position binding, or the prefix-axis information VPG targets, is corrupted. Building a weak-condition branch by reorganizing the model's own activations to avoid additional post-training is a recurring pattern in inference-time guidance~\citep{karras2024autoguidance,ahn2024pag,hong2024seg,softcfg}. Sec.~\ref{sec:experiments-var-ablation} ablates each component of this construction.

Formally, each prefix segment $j\in\{1,\ldots,k-1\}$ uses the downsampled cumulative feature at the corresponding scale. It is embedded with the model's visual projection and scale-position encoding.
\begin{equation}
  e_{j,u} = \mathrm{Emb}(\bar{F}_j[u]) + \mathrm{PosEmb}(j,u),
  \qquad 1\le j<k,
  \label{eq:prefix-embedding}
\end{equation}

where $u$ indexes spatial sites. At each guided step, $\mathcal{S}_k$ is a uniformly random subset of prefix sites of size $|\mathcal{S}_k|=n_p\sum_{j<k}h_j w_j$. For selected pairs $(j,u)\in\mathcal{S}_k$, we sample a donor $\pi(j,u)=(j,u')$ from the same scale $j$ and replace the full embedding.
\begin{equation}
  \tilde e_{j,u}
  =
  \begin{cases}
    e_{j,u'}, & (j,u)\in\mathcal{S}_k,\quad (j,u')=\pi(j,u), \\
    e_{j,u}, & (j,u)\notin\mathcal{S}_k.
  \end{cases}
  \label{eq:prefix-substitution-var}
\end{equation}
The corrupted-prefix prediction $p_{c,\tilde r_{<k}}(r_k)$ uses the replaced embeddings from Eq.~\eqref{eq:prefix-substitution-var}.


\section{Experiments}
\label{sec:experiments}

Our experiments cover class-conditional image generation with VAR~\citep{var}, text-to-image generation with Infinity~\citep{infinity}, and text-to-video generation with InfinityStar~\citep{InfinityStar}.
We first report main results in these three settings, then ablate the corrupted-prefix construction and VPG hyperparameters on VAR.
\subsection{Experimental setup}
\label{sec:experiments-setup}

We evaluate released checkpoints without retraining or architectural changes.
For class-conditional image generation, we use official VAR checkpoints at depths 16/20/24/30 and follow the standard ImageNet~\citep{deng2009imagenet} 256$\times$256 protocol with 50{,}000 generated samples.
We use the ADM evaluation suite and report Fréchet Inception Distance (FID) as the main metric.
For text-to-image generation, we evaluate the released Infinity checkpoint on GenEval~\citep{ghosh2023geneval} and DPG-Bench~\citep{hu2024ella}.
For text-to-video generation, we evaluate the released InfinityStar checkpoint on all 946 VBench prompts~\citep{huang2024vbench}.

\subsection{Main results}
\label{sec:experiments-main}


\paragraph{Class-conditional generation}
\label{sec:experiments-var-scaling}

For VAR, baseline sampling follows the public evaluation scripts with top-$k{=}900$, top-$p{=}0.96$, and CFG weight $1.5$.
VPG uses same-scale full-embedding replacement to construct the corrupted prefix: a fraction $n_p$ of prefix sites is replaced by the full embedding of a uniformly sampled donor site from the same scale, including visual code and scale-position encoding (Sec.~\ref{sec:method-prefix-substitution}).
We set $n_p{=}0.1$ for all VAR depths and select $\lambda$ by best FID for each model size.

\newcommand{\varfid}[1]{\makebox[2.2em][l]{#1}}
\newcommand{\varfidgain}[1]{\makebox[3.0em][l]{\gain{#1}}}
\begin{table}[t]
  \centering
  \caption{\textbf{ImageNet 256$\times$256 class-conditional generation with VAR.}
  $\lambda$ denotes the VPG guidance scale selected by best FID; \gain{green} values report $\Delta$FID relative to the unguided baseline.
  VPG improves FID across all evaluated sizes.}
  \label{tab:var-scaling}
  \small
  \setlength{\tabcolsep}{5pt}
  \renewcommand{\arraystretch}{1.1}
  \begin{tabular}{lccl@{\hspace{12pt}}llll}
    \toprule
    Model & \#Params & $\lambda$ & FID$\downarrow$ &
    Model & \#Params & $\lambda$ & FID$\downarrow$ \\
    \midrule
    VAR-d16            & 310M & ---  & \varfid{3.35}
      & VAR-d24            & 1.0B & ---  & \varfid{2.15} \\
    \quad \textbf{+ VPG (Ours)} & 310M & 3.0  & \varfid{\textbf{2.72}}\,\varfidgain{(-0.63)}
      & \quad \textbf{+ VPG (Ours)} & 1.0B & 1.8  & \varfid{\textbf{1.83}}\,\varfidgain{(-0.32)} \\
    \midrule
    VAR-d20            & 600M & ---  & \varfid{2.67}
      & VAR-d30            & 2.0B & ---  & \varfid{1.94} \\
    \quad \textbf{+ VPG (Ours)} & 600M & 2.4  & \varfid{\textbf{2.28}}\,\varfidgain{(-0.39)}
      & \quad \textbf{+ VPG (Ours)} & 2.0B & 1.3  & \varfid{\textbf{1.84}}\,\varfidgain{(-0.10)} \\
    \bottomrule
  \end{tabular}
\end{table}

\paragraph{Text-to-image generation}
\label{sec:experiments-infinity}

\begin{table}[t]
  \caption{\textbf{Text-to-image evaluation on GenEval and DPG-Bench.}
  $\dagger$ indicates prompt rewriting; \textbf{bold} and \underline{underlined} mark best and second-best results; \gain{green} values report improvements over Infinity.
  VPG ties the best GenEval score, and improves DPG-Bench Overall.}
  \label{tab:infinity-geneval-dpg}
  \centering
  \footnotesize
  \setlength{\tabcolsep}{3pt}
  \begin{tabular}{lc cccc c c}
    \toprule
    & & \multicolumn{4}{c}{GenEval $\uparrow$} & & DPG-Bench $\uparrow$ \\
    \cmidrule(lr){3-6} \cmidrule(lr){8-8}
    Method & \#Params & Two Object & Position & Color Attribute & Overall & & Overall \\
    \midrule
    \rowcolor{gray!15}
    \multicolumn{8}{l}{\emph{Diffusion Models}} \\
    SDv2.1         & 0.9B & 0.51 & 0.07 & 0.17 & 0.50           & & 68.09 \\
    SDXL           & 2.6B & 0.74 & 0.15 & 0.23 & 0.55           & & 74.65 \\
    PixArt-$\Sigma$& 0.6B & 0.62 & 0.14 & 0.27 & 0.55           & & 80.54 \\
    SD3 ($d{=}24$) & 2B   & 0.74 & 0.34 & 0.36 & 0.62           & & \textbf{84.08} \\
    SD3 ($d{=}38$) & 8B   & \textbf{0.89} & 0.34 & 0.47 & \textbf{0.71} & & --    \\
    \midrule
    \rowcolor{gray!15}
    \multicolumn{8}{l}{\emph{Autoregressive Models}} \\
    LlamaGen       & 0.8B & 0.34 & 0.07 & 0.04 & 0.32              & & --    \\
    Show-o         & 1.3B & 0.80 & 0.31 & \underline{0.50} & 0.68              & & 67.48 \\
    Emu3           & 8.5B & 0.81$^\dagger$ & \textbf{0.49}$^\dagger$ & 0.45$^\dagger$ & 0.66$^\dagger$ & & 81.60 \\
    Infinity       & 2B   & 0.83$^\dagger$ & 0.39$^\dagger$ & \textbf{0.56}$^\dagger$ & \underline{0.70}$^\dagger$ & & 83.46 \\
    \quad\textbf{+ VPG (Ours)} & 2B 
    & \makecell[c]{\underline{0.85}$^\dagger$\\[-1pt]\scriptsize\gain{(+0.02)}} 
    & \makecell[c]{\underline{0.41}$^\dagger$\\[-1pt]\scriptsize\gain{(+0.02)}} 
    & \makecell[c]{\textbf{0.56}$^\dagger$\\[-1pt]\scriptsize\gain{(+0.00)}} 
    & \makecell[c]{\textbf{0.71}$^\dagger$\\[-1pt]\scriptsize\gain{(+0.01)}} 
    & 
    & \makecell[c]{\underline{83.80}\\[-1pt]\scriptsize\gain{(+0.34)}} \\
    \bottomrule
  \end{tabular}
  \vspace{-1.0em}
\end{table}

Tab.~\ref{tab:var-scaling} reports the best-FID operating point of VPG at each VAR depth.
Across all four capacities, VPG lowers FID.
The average reduction is 0.36 FID, with the largest gain on VAR-d16 (3.35 $\rightarrow$ 2.72).
The most competitive operating point is VAR-d24+VPG: its FID of 1.83 is slightly better than the unguided VAR-d30 (1.94) and nearly identical to VAR-d30+VPG (1.84), despite using half the parameters.
The gain becomes smaller as model capacity increases, suggesting that prefix contrast is most useful when the generator has less capacity.


Tab.~\ref{tab:infinity-geneval-dpg} reproduces the Infinity comparison table~\citep{infinity} and includes our VPG result. 
We follow the same sampling pipeline as the Infinity baselines and apply same-scale full-embedding replacement with $n_p{=}0.1$ and $\lambda{=}0.2$. 
\textbf{Infinity + VPG} improves the more compositional GenEval dimensions, including Two Obj. and Position, by +0.02 each, while maintaining Infinity's best Color Attribute score of 0.56. 
As a result, the GenEval Overall score increases from 0.70 to 0.71, matching the best score in the table and achieving the strongest performance among autoregressive models.
On DPG-Bench, VPG improves Infinity from 83.46 to 83.80.

\paragraph{Text-to-video generation}
\label{sec:experiments-infinitystar}

Tab.~\ref{tab:infinitystar-vbench} compares the released InfinityStar checkpoint with and without VPG under the $\dagger$ protocol. 
We apply same-scale full-embedding replacement with $n_p{=}0.05$ and $\lambda{=}0.25$. 
A smaller corruption fraction is important for video generation, as long spatiotemporal rollouts are more sensitive to prefix perturbations; see Appendix~\ref{app:infinitystar-schedule} for more details. 
VPG improves InfinityStar's Overall score from 83.86 to 84.35. 
The most notable gains are semantic: VPG achieves the best Multi-object score of 89.63 and the best Semantic score of 83.51 among all listed models, while also improving Scene to the second-best result.

\begin{table}[t]
  \caption[Text-to-video evaluation on VBench]{\textbf{Text-to-video evaluation on VBench.} \textbf{bold} and \underline{underlined} mark best and second-best results; \gain{green} values report changes relative to unguided InfinityStar.
  VPG gives InfinityStar the best Multi-object and Semantic scores while improving Overall.}
  \label{tab:infinitystar-vbench}
  \centering
  \footnotesize
  \setlength{\tabcolsep}{3pt}
  \begin{tabular}{lcccccccc}
    \toprule
    Model & \#Params 
    & \makecell{Human\\Action}
    & Scene
    & \makecell{Multi\\object}
    & \makecell{Appearance\\Style}
    & \makecell{Quality\\Score}
    & \makecell{Semantic\\Score}
    & \makecell{Overall} \\
    \midrule
    \rowcolor{gray!15}
    \multicolumn{9}{l}{\emph{Diffusion Models}} \\
    AnimateDiff-V2   & 1.5B & 92.60 & 50.19 & 36.88 & 22.42 & 82.90 & 69.75 & 80.27 \\
    VideoCrafter-2.0 & 1.5B & 95.00 & 55.29 & 40.66 & \textbf{25.13} & 82.20 & 73.42 & 80.44 \\
    OpenSora V1.2    & 1.1B & 85.80 & 42.47 & 58.41 & 23.89 & 80.71 & 73.30 & 79.23 \\
    Show-1           & 6B   & 95.60 & 47.03 & 45.47 & 23.06 & 80.42 & 72.98 & 78.93 \\
    Gen-3            & --   & 96.40 & 54.57 & 53.64 & 24.31 & 84.11 & 75.17 & 82.32 \\
    CogVideoX-5B     & 5B   & \textbf{99.40} & 53.20 & 62.11 & \underline{24.91} & 82.75 & 77.04 & 81.61 \\
    HunyuanVideo     & 13B  & 94.40 & 53.88 & 68.55 & 19.80 & 85.09 & 75.82 & 83.24 \\
    Goku             & 2B   & 97.60 & \textbf{57.08} & 79.48 & 23.08 & \underline{85.60} & 81.87 & \textbf{84.85} \\
    Wan~2.1          & 14B  & \underline{98.80} & 53.67 & 81.44 & 21.13 & \textbf{85.64} & 80.95 & \underline{84.70} \\
    \midrule
    \rowcolor{gray!15}
    \multicolumn{9}{l}{\emph{Autoregressive Models}} \\
    Nova             & 0.6B & 95.20 & 54.06 & 77.52 & 20.92 & 80.39 & 79.05 & 80.12 \\
    Emu3             & 8B   & 77.71 & 37.11 & 44.64 & 20.92 & 84.09 & 68.43 & 80.96 \\
    InfinityStar     & 8B   & 98.00 & 54.51 & \underline{87.50} & 20.54 & 84.14 & \underline{82.74} & 83.86 \\
    \quad\textbf{+ VPG (Ours)}
    & 8B
    & \makecell{98.00\\[-1pt]\scriptsize\gain{(+0.00)}}
    & \makecell{\underline{56.61}\\[-1pt]\scriptsize\gain{(+2.10)}}
    & \makecell{\textbf{89.63}\\[-1pt]\scriptsize\gain{(+2.13)}}
    & \makecell{20.60\\[-1pt]\scriptsize\gain{(+0.06)}}
    & \makecell{84.56\\[-1pt]\scriptsize\gain{(+0.42)}}
    & \makecell{\textbf{83.51}\\[-1pt]\scriptsize\gain{(+0.77)}}
    & \makecell{84.35\\[-1pt]\scriptsize\gain{(+0.49)}} \\
    \bottomrule
  \end{tabular}
  \vspace{-1.0em}
\end{table}

\subsection{Qualitative study}
\label{sec:experiments-qualitative}

\begin{figure}[!t]
  \centering
  \includegraphics[width=0.9\linewidth]{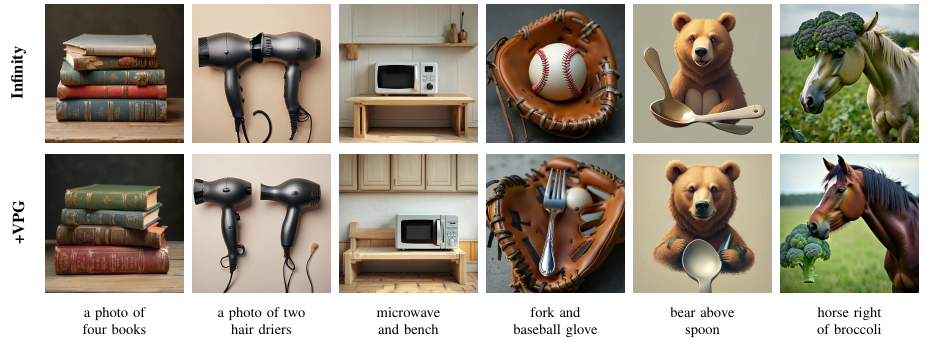}
  \caption{\textbf{GenEval qualitative comparison for Infinity vs. Infinity + VPG.} VPG corrects failures in counting, two-object binding, and spatial position.}
  \label{fig:geneval-vis-grid}
  \vspace{-1.0em}
\end{figure}

Fig.~\ref{fig:geneval-vis-grid} shows matched GenEval examples from categories where Infinity + VPG improves over the Infinity baseline. The examples illustrate the same trend as Tab.~\ref{tab:infinity-geneval-dpg}: VPG helps when the unguided autoregressive sampler drops objects, merges two entities, or loses spatial relations under a strong scene prior. Additional visualizations are provided in Appendix~\ref{app:qualitative}, including class-conditional VAR comparisons (Figs.~\ref{fig:var-qualitative-golden}--\ref{fig:var-qualitative-cheeseburger}), InfinityStar video-frame comparisons (Figs.~\ref{fig:infinitystar-train-library}--\ref{fig:infinitystar-paper-whale-harbor}), and DPG-Bench text-to-image examples (Figs.~\ref{fig:dpg-vis-01}--\ref{fig:dpg-vis-19}).

\subsection{Ablations}
\label{sec:experiments-ablations}

We ablate VPG on class-conditional VAR because it provides a controlled protocol with 50k-sample FID evaluation.
The ablations answer two questions.
(i) what kind of corrupted prefix creates a useful weak reference?
(ii) how sensitive is VPG to the guidance scale $\lambda$ and corruption fraction $n_p$?
All ablations use the ImageNet 256$\times$256 ADM protocol, making them directly comparable to Tab.~\ref{tab:var-scaling}.

\paragraph{Effect of corrupted-prefix construction}
\label{sec:experiments-var-ablation}

Tab.~\ref{tab:ablation-corruption} ablates how the corrupted prefix $\tilde r_{<k}$ is built on VAR-d16 with fixed $n_p{=}0.1$.
The replacement rules are formalized in Appendix~\ref{app:vpg-replacement-variants}.
The three incomplete or off-manifold corruptions all increase FID relative to the 3.35 baseline: random codebook gives $+3.91$ (7.26), same-scale token gives $+2.52$ (5.87), and same-scale position gives $+1.11$ (4.46).
Only same-scale full-embedding replacement improves over the baseline, reducing FID by $0.63$ (3.35 $\rightarrow$ 2.72), a relative reduction of $18.8\%$.
This supports the design criterion in Sec.~\ref{sec:method-prefix-substitution}: the corrupted branch must be weak enough to break content-position binding, but close enough to preserve same-scale input statistics.
The corresponding FID/IS curves are provided in Appendix~\ref{app:vpg-replacement-variants}.

\begin{wraptable}{r}{0.42\linewidth}
  \vspace{-\baselineskip}
  \centering
  \caption{\textbf{Corrupted-prefix replacement variants on VAR-d16.}
  each row reports the best FID for one replacement strategy ($\lambda>0$ ).
  Same-scale embed. is the only variant that improves over the baseline.}
  \label{tab:ablation-corruption}
  \small
  \setlength{\tabcolsep}{4pt}
  \begin{tabular}{ll}
    \toprule
    Variant & Best FID$\downarrow$ \\
    \midrule
    None (baseline)                     & 3.35 \\
    Random codebook                     & 7.26 \\
    Same-scale token                    & 5.87 \\
    Same-scale position                 & 4.46 \\
    \textbf{Same-scale embed.\ (Ours)}  & \textbf{2.72}\,\gain{(-0.63)} \\
    \bottomrule
  \end{tabular}
  \vspace{-2.4\baselineskip}
\end{wraptable}
\paragraph{Sensitivity to guidance strength and corruption fraction}
\label{sec:experiments-var-sensitivity}

\begin{figure}[t]
  \centering
  \begin{minipage}[t]{0.48\linewidth}
    \centering
    \includegraphics[width=\linewidth]{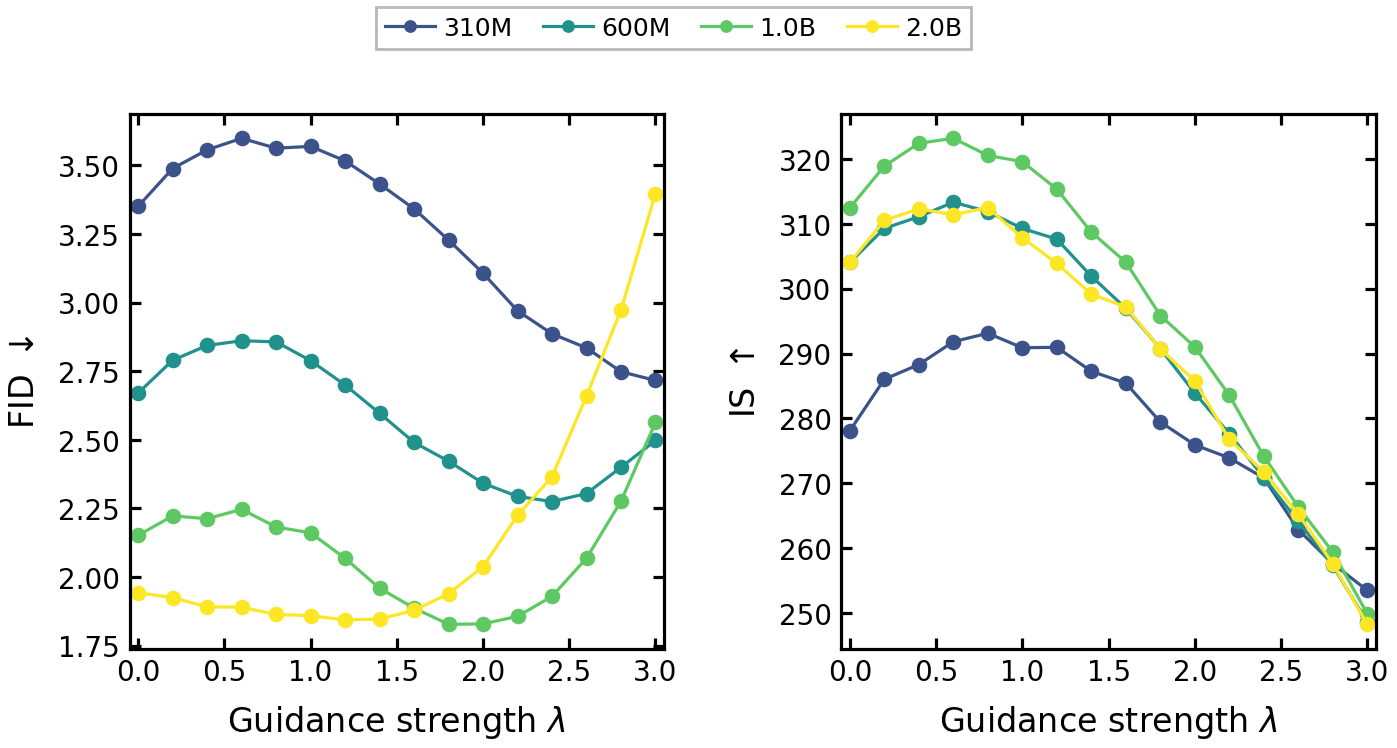}\\[0.5ex]
    {\small \textbf{(Left)} FID, IS vs. $\lambda$ ($n_p{=}0.1$, across model sizes).}
  \end{minipage}%
  \hfill
  \begin{minipage}[t]{0.48\linewidth}
    \centering
    \includegraphics[width=\linewidth]{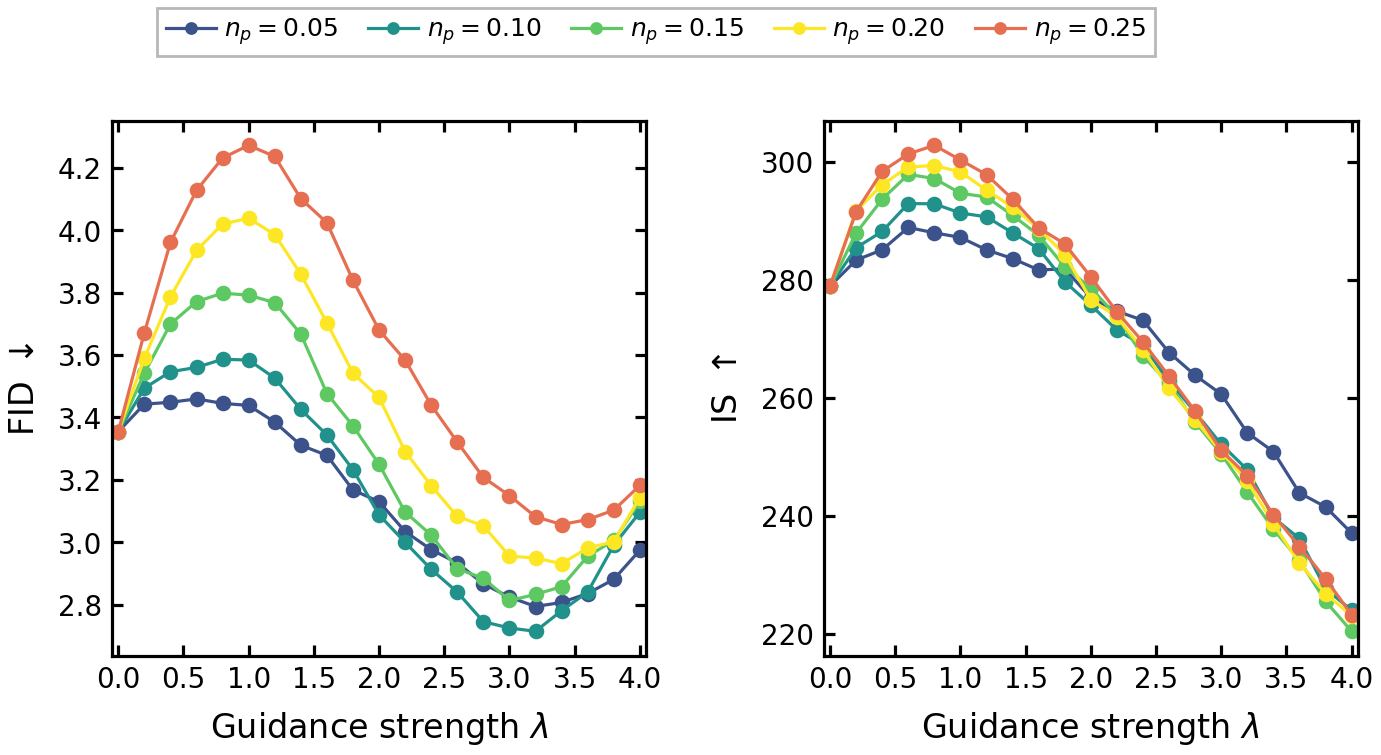}\\[0.5ex]
    {\small \textbf{(Right)} FID, IS vs. $\lambda$ for corruption levels.}
  \end{minipage}
  \caption{\textbf{VPG sweeps on class-conditional VAR (ImageNet 256$\times$256).}
  \textbf{Left:} guidance strength $\lambda$ with fixed corruption fraction $n_p{=}0.1$, across four VAR capacities.
  \textbf{Right:} the same $\lambda$ sweep on VAR-d16 with the corruption fraction varied over $n_p\in\{0.05,0.10,0.15,0.20,0.25\}$; larger $n_p$ amplifies the FID/IS response to $\lambda$ while small $n_p$ remains close to the baseline.}
  \label{fig:var-vpg-lambda-np}
  \vspace{-1em}
\end{figure}

Fig.~\ref{fig:var-vpg-lambda-np} studies the two main inference hyperparameters.
The left panel varies the guidance strength $\lambda$ across four VAR capacities with fixed $n_p{=}0.1$.
VPG consistently reaches a lower FID than the unguided baseline, while the best $\lambda$ decreases as model size increases.
The right panel fixes VAR-d16 and sweeps $n_p\in\{0.05,0.10,0.15,0.20,0.25\}$ across the same $\lambda$ range.
Larger corruption fractions amplify the response to $\lambda$, whereas $n_p{=}0.05$ remains close to the baseline.
Across the tested range, moderate corruption and a sufficiently large guidance scale give the best FID trade-off, matching the operating point used in Tab.~\ref{tab:var-scaling}.

\section{Conclusion}
\label{sec:conclusion}

We presented Visual Prefix Guidance (VPG), a training-free sampling rule that treats the generated prefix as an explicit guidance source in visual autoregressive generation. VPG favors next-step candidates that better support the existing prefix by contrasting genuine-prefix and corrupted-prefix predictions, yielding a simple logit extrapolation without auxiliary models or retraining.

VPG complements CFG: CFG sharpens dependence on the external condition, while VPG sharpens dependence on the internal visual history. Across VAR, Infinity, and InfinityStar, this prefix-axis guidance improves image and video generation, especially when the unguided sampler drops objects, weakens spatial relations, or lets video semantics drift. These results show that prefix support is a practical inference-time control signal for autoregressive visual generation.

{
\small
\bibliographystyle{plainnat}
\bibliography{main}
}

\clearpage
\appendix

\section{Theoretical derivations}
\label{app:cfg-vpg-derivation}

This section gives the full compatibility-augmentation derivations for CFG (Sec.~\ref{sec:prelim-ar-cfg}) and VPG (Sec.~\ref{sec:method-vpg}) in next-scale visual autoregression. Both methods are obtained by exponentiating a posterior compatibility term in an augmented (unnormalized) conditional, then substituting a Bayes identity that expresses the compatibility as a log-ratio between two next-scale predictions of the same frozen model. The two derivations differ only in which axis is contrasted: the external condition $c$ for CFG, and the generated prefix $r_{<k}$ versus a corrupted prefix $\tilde r_{<k}$ for VPG.

\subsection{\texorpdfstring{CFG: enhancing $p(c\mid r_{\le k})$}{CFG: enhancing p(c | r\_{<=k})}}
\label{app:cfg-derivation}

\paragraph{Augmented conditional.}
We start from $p_\theta(r_k\mid r_{<k}, c)$ and define the augmented (unnormalized) conditional by exponentiating the posterior compatibility $p(c\mid r_{\le k})$, where $r_{\le k}=(r_{<k},r_k)$:
\begin{equation}
  p_\theta^{\mathrm{CFG}}(r_k\mid r_{<k}, c)
  \;\propto\;
  p_\theta(r_k\mid r_{<k}, c)\,p(c\mid r_{\le k})^{\gamma},
  \label{eq:app-cfg-augmented}
\end{equation}
with guidance strength $\gamma\ge 0$. Taking $\log$ gives, up to a constant in $r_k$,
\begin{equation}
  \log p_\theta^{\mathrm{CFG}}(r_k\mid r_{<k}, c)
  \;=\;
  \log p_\theta(r_k\mid r_{<k}, c) + \gamma\,\log p(c\mid r_{\le k}) + \mathrm{const}.
  \label{eq:app-cfg-aug-log}
\end{equation}

\paragraph{Bayes decomposition along the AR chain.}
Applying Bayes' rule to the prefix-conditioned joint $p(r_k, c\mid r_{<k})$ yields
\begin{equation}
  p_\theta(r_k\mid r_{<k}, c)
  \;=\;
  \frac{p(c\mid r_{\le k})\,p_\theta(r_k\mid r_{<k})}{p(c\mid r_{<k})}.
  \label{eq:app-cfg-bayes}
\end{equation}
Since $p(c\mid r_{<k})$ does not depend on $r_k$, taking $\log$ and rearranging Eq.~\eqref{eq:app-cfg-bayes} gives, up to a constant in $r_k$,
\begin{equation}
  \log p(c\mid r_{\le k})
  \;=\;
  \log p_\theta(r_k\mid r_{<k}, c) - \log p_\theta(r_k\mid r_{<k}) + \mathrm{const}.
  \label{eq:app-cfg-classifier}
\end{equation}
Eq.~\eqref{eq:app-cfg-classifier} is the AR analogue of the classifier-gradient identity used by diffusion CFG: the implicit ``classifier'' that scores how well the prefix-and-current-token sequence supports $c$ is exactly the log-ratio of the conditional and unconditional next-scale distributions.

Substituting this identity back into Eq.~\eqref{eq:app-cfg-augmented} gives the distribution-level form used in the main text:
\begin{equation}
  p_\theta^{\mathrm{CFG}}(r_k\mid r_{<k}, c)
  \;\propto\;
  p_\theta(r_k\mid r_{<k}, c)
  \left(
    \frac{p_\theta(r_k\mid r_{<k}, c)}
         {p(r_k\mid r_{<k})}
  \right)^\gamma.
  \label{eq:app-cfg-ratio-distribution}
\end{equation}
Thus the classifier posterior $p(c\mid r_{\le k})^\gamma$ is never evaluated by a separate classifier; it is represented by a likelihood ratio between two next-scale predictions.

\paragraph{From two predictions to guided logits.}
CFG approximates the unconditional $p(r_k\mid r_{<k})$ with the same generative model evaluated under a learned null condition $\emptyset$, $p(r_k\mid r_{<k})\approx p_\theta(r_k\mid r_{<k}, \emptyset)$. Substituting Eq.~\eqref{eq:app-cfg-classifier} into Eq.~\eqref{eq:app-cfg-aug-log} expresses the augmented log-density as a linear combination of the two predictions:
\begin{equation}
  \log p_\theta^{\mathrm{CFG}}(r_k\mid r_{<k}, c)
  \;=\;
  (1+\gamma)\,\log p_\theta(r_k\mid r_{<k}, c)
  - \gamma\,\log p_\theta(r_k\mid r_{<k}, \emptyset)
  + \mathrm{const}.
  \label{eq:app-cfg-distribution}
\end{equation}
Because visual autoregressive models output logits before the softmax, Eq.~\eqref{eq:app-cfg-distribution} translates directly to the logit-space extrapolation rule in Eq.~\eqref{eq:cfg-logits}. With $\ell_k^c\equiv\ell_k(c,r_{<k})$ and $\ell_k^\emptyset\equiv\ell_k(\emptyset,r_{<k})$,
\begin{equation*}
  \ell^{\mathrm{CFG}}_{k}
  \;=\;
  (1+\gamma)\,\ell_k^c
  - \gamma\,\ell_k^\emptyset.
\end{equation*}
\textit{Remark (parameterization).} Many papers write CFG with a guidance scale $s\ge 1$ as $\ell^{\mathrm{CFG}}_k=\ell_k^\emptyset+s\,(\ell_k^c-\ell_k^\emptyset)$, which matches Eq.~\eqref{eq:cfg-logits} by setting $s=1+\gamma$.

\subsection{\texorpdfstring{VPG: enhancing $p(r_{<k}\mid r_k, c)$}{VPG: enhancing p(prefix | r\_k, c)}}
\label{app:vpg-derivation}

VPG applies the same compatibility-augmentation logic along the prefix axis. Whereas CFG enhances $p(c\mid r_{\le k})$ by contrasting the external condition $c$ against the null condition $\emptyset$ at fixed prefix, VPG enhances the posterior of the prefix given the next-scale token map and the external condition, $p(r_{<k}\mid r_k, c)$, by contrasting the genuine generated prefix $r_{<k}$ against an inference-time corrupted prefix $\tilde r_{<k}$ while holding $c$ fixed.

\paragraph{Augmented conditional.}
We start from $p_\theta(r_k\mid r_{<k}, c)$ and define the augmented (unnormalized) conditional by exponentiating the posterior compatibility $p(r_{<k}\mid r_k, c)$:
\begin{equation}
  p_\theta^{\mathrm{VPG}}(r_k\mid r_{<k}, c)
  \;\propto\;
  p_\theta(r_k\mid r_{<k}, c)\,p(r_{<k}\mid r_k, c)^{\lambda},
  \label{eq:app-vpg-augmented}
\end{equation}
with prefix-guidance strength $\lambda\ge 0$. Taking $\log$ gives, up to a constant in $r_k$,
\begin{equation}
  \log p_\theta^{\mathrm{VPG}}(r_k\mid r_{<k}, c)
  \;=\;
  \log p_\theta(r_k\mid r_{<k}, c) + \lambda\,\log p(r_{<k}\mid r_k, c) + \mathrm{const}.
  \label{eq:app-vpg-aug-log}
\end{equation}

\paragraph{Bayes decomposition along the AR chain.}
Applying Bayes' rule to the prefix-axis posterior yields
\begin{equation}
  p(r_{<k}\mid r_k, c)
  \;=\;
  \frac{p_\theta(r_k\mid r_{<k}, c)\, p(r_{<k}\mid c)}{p(r_k\mid c)},
  \label{eq:app-vpg-bayes}
\end{equation}
where the marginal $p(r_k\mid c) = \int p_\theta(r_k\mid r_{<k}, c)\, p(r_{<k}\mid c)\,\mathrm{d}\mu(r_{<k})$ is taken over prefixes drawn from the model's prefix distribution, with $\mu$ denoting the appropriate base measure. For discrete token prefixes, this integral is a sum under the counting measure. Since $p(r_{<k}\mid c)$ does not depend on $r_k$, taking $\log$ and rearranging Eq.~\eqref{eq:app-vpg-bayes} gives, up to a constant in $r_k$,
\begin{equation}
  \log p(r_{<k}\mid r_k, c)
  \;=\;
  \underbrace{\log p_\theta(r_k\mid r_{<k}, c)}_{\text{prefix-conditional}}
  -
  \underbrace{\log p(r_k\mid c)}_{\text{prefix-marginalized}}
  + \mathrm{const}.
  \label{eq:app-vpg-classifier}
\end{equation}
Eq.~\eqref{eq:app-vpg-classifier} is the prefix-axis analogue of Eq.~\eqref{eq:app-cfg-classifier}: the implicit ``classifier'' that scores how well the next-scale token map $r_k$ supports the genuine prefix $r_{<k}$ is the log-ratio between the prefix-conditional model and the prefix-marginalized predictive.

Substituting this identity back into Eq.~\eqref{eq:app-vpg-augmented} gives the distribution-level form used in the main text:
\begin{equation}
  p_\theta^{\mathrm{VPG}}(r_k\mid r_{<k}, c)
  \;\propto\;
  p_\theta(r_k\mid r_{<k}, c)
  \left(
    \frac{p_\theta(r_k\mid r_{<k}, c)}
         {p(r_k\mid c)}
  \right)^\lambda.
  \label{eq:app-vpg-ratio-distribution}
\end{equation}
Thus the prefix posterior $p(r_{<k}\mid r_k,c)^\lambda$ is never evaluated by a separate classifier; it is represented by a likelihood ratio between the prefix-conditional next-scale prediction and the prefix-marginalized predictive.

\begin{wrapfigure}{r}{0.48\textwidth}
    \vspace{-1.2em}
    \centering
    \includegraphics[width=0.46\textwidth]{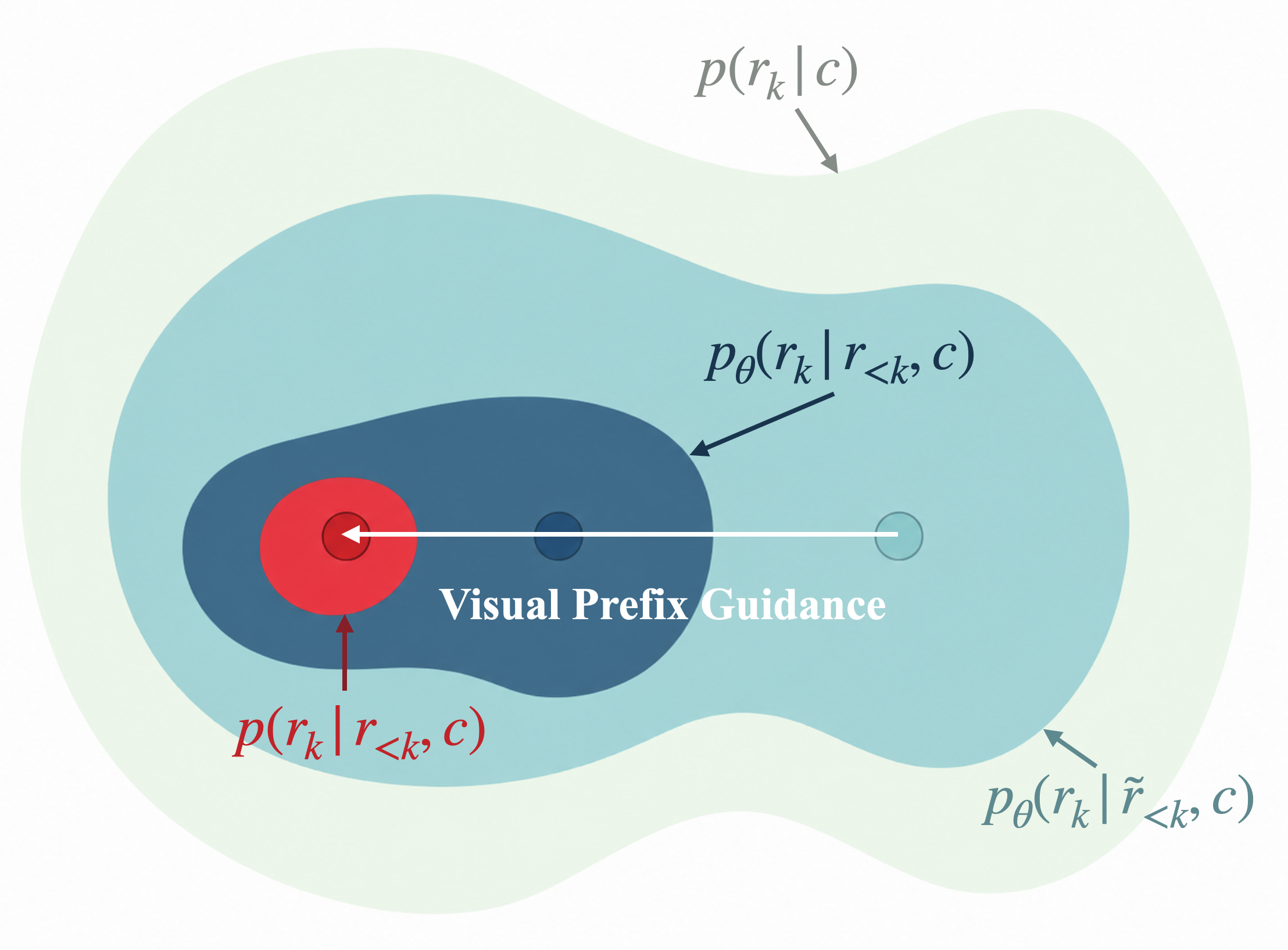}
    \caption{\small Illustration of Visual Prefix Guidance.}
    \label{fig:corrupted-prefix-surrogate}
\end{wrapfigure}

\paragraph{Corrupted-prefix surrogate for the prefix-marginalized predictive.}
The prefix-marginalized predictive distribution $p_\theta(r_k\mid c)$ is not directly accessible from a frozen visual AR model. 
In principle, it requires marginalizing over all possible prefixes, which is intractable. 
This is analogous to CFG, where the unconditional branch cannot be obtained from a conditional model unless the model has been trained with a null-condition mode. CFG addresses this issue by using a learned null condition, i.e.,
\begin{equation}
    p_\theta(r_k\mid r_{<k})
    \approx
    p_\theta(r_k\mid r_{<k},\emptyset).
\end{equation}

For visual AR models, one could similarly train a dedicated null-prefix mode for next-token prediction, but this would require retraining the model. 
Instead, we seek an inference-time surrogate prefix that satisfies three desiderata: 
(i) it should provide a weaker reference than the clean-prefix prediction $p_\theta(r_k\mid r_{<k},c)$; 
(ii) it should better reflect the role of the prefix-marginalized predictive $p_\theta(r_k\mid c)$ in the guidance objective; 
and (iii) it should remain within, or at least close to, the training distribution of the frozen model.

A simple surrogate is to use a uniformly random token prefix as a null-prefix condition. 
Let
\begin{equation}
    U_{<k}=(U_1,\ldots,U_{k-1}),
    \qquad
    U_i \overset{\mathrm{i.i.d.}}{\sim} \mathrm{Unif}(\mathcal{V}),
\end{equation}
where $\mathcal{V}$ denotes the visual token vocabulary. 
The corresponding random-prefix reference can be written as
\begin{equation}
    p_{\textnormal{ref}}^{\mathrm{rand}}(r_k\mid c)
    :=
    p_\theta(r_k\mid U_{<k},c).
    \label{eq:app-vpg-uniform}
\end{equation}
However, since the frozen AR model has not been trained to condition on such random prefixes, this input can fall outside the training distribution and may lead to degenerate or unstable guidance. 
This motivates using a perturbed version of the observed visual prefix rather than an independently sampled random prefix since the observed visual prefix falls into the training distribution.

VPG therefore constructs the reference along the autoregressive prefix axis by evaluating the same frozen model under an inference-time corrupted prefix $\tilde r_{<k}$:
\begin{equation}
    p_{\textnormal{ref}}(r_k\mid c)
    :=
    p_\theta(r_k\mid \tilde r_{<k},c).
    \label{eq:app-vpg-surrogate}
\end{equation}
The corrupted prefix has the same scale and format as the clean prefix, allowing the frozen transformer to process it in a familiar input regime, while carrying weaker prefix-specific evidence than $r_{<k}$. 
This reference can be viewed as an interpolation between the clean-prefix prediction $p_\theta(r_k\mid r_{<k},c)$ and the ideal prefix-marginalized predictive $p_\theta(r_k\mid c)$, balancing input validity against approximation accuracy.
It therefore serves as a weak-prefix reference, playing a role analogous to the null condition $\emptyset$ in CFG, but along the autoregressive prefix axis rather than the external-condition axis.

\paragraph{From two predictions to guided logits.}
Substituting Eq.~\eqref{eq:app-vpg-surrogate} into Eq.~\eqref{eq:app-vpg-classifier} and then into Eq.~\eqref{eq:app-vpg-aug-log} expresses the augmented log-density as a linear combination of the genuine-prefix and corrupted-prefix predictions:
\begin{equation}
  \log p_\theta^{\mathrm{VPG}}(r_k\mid r_{<k}, c)
  \;=\;
  (1+\lambda)\,\log p_\theta(r_k\mid r_{<k}, c)
  - \lambda\,\log p_\theta(r_k\mid \tilde r_{<k}, c)
  + \mathrm{const}.
  \label{eq:app-vpg-distribution}
\end{equation}
Translating to logit space, Eq.~\eqref{eq:app-vpg-distribution} recovers the VPG extrapolation rule used in Eq.~\eqref{eq:vpg-logits}. Writing $\ell_k^{\mathrm{gen}}\equiv\ell_k(c,r_{<k})$ and $\ell_k^{\mathrm{corr}}\equiv\ell_k(c,\tilde r_{<k})$,
\begin{equation*}
  \ell^{\mathrm{VPG}}_{k}
  \;=\;
  \ell_k^{\mathrm{gen}}
  + \lambda\bigl(\ell_k^{\mathrm{gen}} - \ell_k^{\mathrm{corr}}\bigr)
  \;=\;
  (1+\lambda)\,\ell_k^{\mathrm{gen}}
  - \lambda\,\ell_k^{\mathrm{corr}}.
\end{equation*}

\paragraph{Interpretation.}
The derivation shows VPG is the prefix-axis dual of CFG. The two methods share the same template: an augmented conditional that exponentiates a posterior compatibility term, a Bayes identity that rewrites the compatibility as a log-ratio between a conditional model and a marginal predictive, and a paired-prediction surrogate for the otherwise-inaccessible marginal. The dualities are summarized below.
\begin{center}
\begin{tabular}{lll}
  \toprule
  & CFG & VPG \\
  \midrule
  Compatibility enhanced & $p(c\mid r_{\le k})$ & $p(r_{<k}\mid r_k, c)$ \\
  Required marginal     & $p(r_k\mid r_{<k})$  & $p(r_k\mid c)$ \\
  Surrogate             & null condition $\emptyset$ & corrupted prefix $\tilde r_{<k}$ \\
  Fixed axis            & generated prefix $r_{<k}$ & external condition $c$ \\
  \bottomrule
\end{tabular}
\end{center}
Both are paired-prediction methods that exploit the same Bayes identity to avoid an explicit auxiliary classifier and require no retraining of the frozen autoregressive model.

\subsection{Composing CFG and VPG}
\label{app:cfg-vpg-composition}

The implementation composes the two guidance operations sequentially. First, CFG is applied at each prefix branch. For an arbitrary prefix branch $a_{<k}$,
\begin{equation}
  p_\theta^{\mathrm{CFG}}(r_k\mid a_{<k}, c)
  \;\propto\;
  p_\theta(r_k\mid a_{<k}, c)\,
  p(c\mid a_{<k}, r_k)^\gamma.
  \label{eq:app-branch-cfg}
\end{equation}
Using the CFG derivation in Sec.~\ref{app:cfg-derivation}, Eq.~\eqref{eq:app-branch-cfg} is implemented by the branch-wise CFG logits
\[
  g_k(a_{<k})
  =
  \ell_k(c,a_{<k})
  + \gamma\bigl(\ell_k(c,a_{<k})-\ell_k(\emptyset,a_{<k})\bigr).
\]
Second, VPG contrasts the CFG-guided distribution under the genuine prefix against the CFG-guided distribution under the corrupted prefix:
\begin{equation}
  p_\theta^{\mathrm{CFG+VPG}}(r_k\mid r_{<k}, c)
  \;\propto\;
  p_\theta^{\mathrm{CFG}}(r_k\mid r_{<k}, c)
  \left(
    \frac{
      p_\theta^{\mathrm{CFG}}(r_k\mid r_{<k}, c)
    }{
      p_\theta^{\mathrm{CFG}}(r_k\mid \tilde r_{<k}, c)
    }
  \right)^\lambda.
  \label{eq:app-cfgvpg-augmented}
\end{equation}
Taking logits of Eq.~\eqref{eq:app-cfgvpg-augmented} directly yields the update in Eq.~\eqref{eq:cfg-vpg-logits}. With $g_k^{\mathrm{gen}}=g_k(r_{<k})$ and $g_k^{\mathrm{corr}}=g_k(\tilde r_{<k})$,
\[
  \ell_k^{\mathrm{CFG+VPG}}
  =
  g_k^{\mathrm{gen}}
  + \lambda\bigl(g_k^{\mathrm{gen}}-g_k^{\mathrm{corr}}\bigr).
\]
This composition differs from exponentiating the original CFG and VPG compatibility terms once in a single raw-model conditional; that alternative would not produce Eq.~\eqref{eq:cfg-vpg-logits}. The sequential form matches the actual sampler: CFG first sharpens the text-condition axis on each branch, and VPG then contrasts the two CFG-guided branches along the prefix axis.

\section{Experimental details and ablations}
\label{app:experimental-details}

This section collects implementation details that support the main experimental claims in Sec.~\ref{sec:experiments}. We first formalize the corrupted-prefix replacement variants used in the VAR ablation, then report the InfinityStar schedule and latency measurements referenced by the text-to-video results.

\subsection{VPG corrupted-prefix replacement variants}
\label{app:vpg-replacement-variants}

Sec.~\ref{sec:experiments-var-ablation} ablates how the corrupted prefix $\tilde r_{<k}$ should be constructed on VAR-d16 (Tab.~\ref{tab:ablation-corruption}). All variants use the same evaluation protocol: the base sampler, checkpoint, ImageNet validation metric pipeline, corruption probability $n_p{=}0.1$, and guidance-strength sweep are fixed; only the replacement rule for selected prefix sites changes. We report each run by its change from the unguided baseline,
\[
\Delta_{\mathrm{FID}}=\mathrm{FID}_{\mathrm{variant}}-\mathrm{FID}_{\mathrm{baseline}},
\qquad
\mathrm{FID}_{\mathrm{baseline}}=3.35.
\]

The ablation tests three design requirements for the corrupted-prefix branch: it should weaken prefix evidence, remain near the model's familiar same-scale input distribution, and disrupt content-position binding rather than only content or only position. We use the same prefix-embedding notation as Sec.~\ref{sec:method-prefix-substitution}: $e_{j,u}=\mathrm{Emb}(\bar F_j[u])+\mathrm{PosEmb}(j,u)$ at scale $j$ and spatial site $u$. For a selected site $(j,u)$, the same-scale variants sample a donor site $u'$ from the same scale; unselected sites remain unchanged.

\paragraph{Random codebook replacement.} Replaces selected sites with embeddings drawn from the global VQ codebook rather than from same-scale prefix activations. This creates a weak reference but can move the corrupted branch off the model's scale-conditioned input manifold, so it tests whether ``weak'' alone is sufficient.

\paragraph{Same-scale token replacement.} Uses
\[
  \tilde e_{j,u}
  =
  \mathrm{Emb}(\bar F_j[u'])
  +
  \mathrm{PosEmb}(j,u).
\]
The visual projection is weakened by copying the donor feature from another same-scale site, but the original scale-position encoding is preserved. This tests whether content weakening alone provides a useful prefix contrast.

\paragraph{Same-scale position replacement.} Uses
\[
  \tilde e_{j,u}
  =
  \mathrm{Emb}(\bar F_j[u])
  +
  \mathrm{PosEmb}(j,u').
\]
The visual content is preserved, but its position-associated embedding is copied from another same-scale site. This tests whether perturbing spatial alignment alone is sufficient.

\paragraph{Same-scale full-embedding replacement (Ours).} Uses
\[
  \tilde e_{j,u}
  =
  \mathrm{Emb}(\bar F_j[u'])
  +
  \mathrm{PosEmb}(j,u')
  =
  e_{j,u'}.
\]
The entire donor embedding is copied from another same-scale site. This preserves same-scale embedding statistics while breaking the binding between local content and its original position, matching the intended weak-prefix surrogate in Sec.~\ref{sec:method-prefix-substitution}.

\begin{figure}[!htbp]
  \centering
  \includegraphics[width=0.8\linewidth]{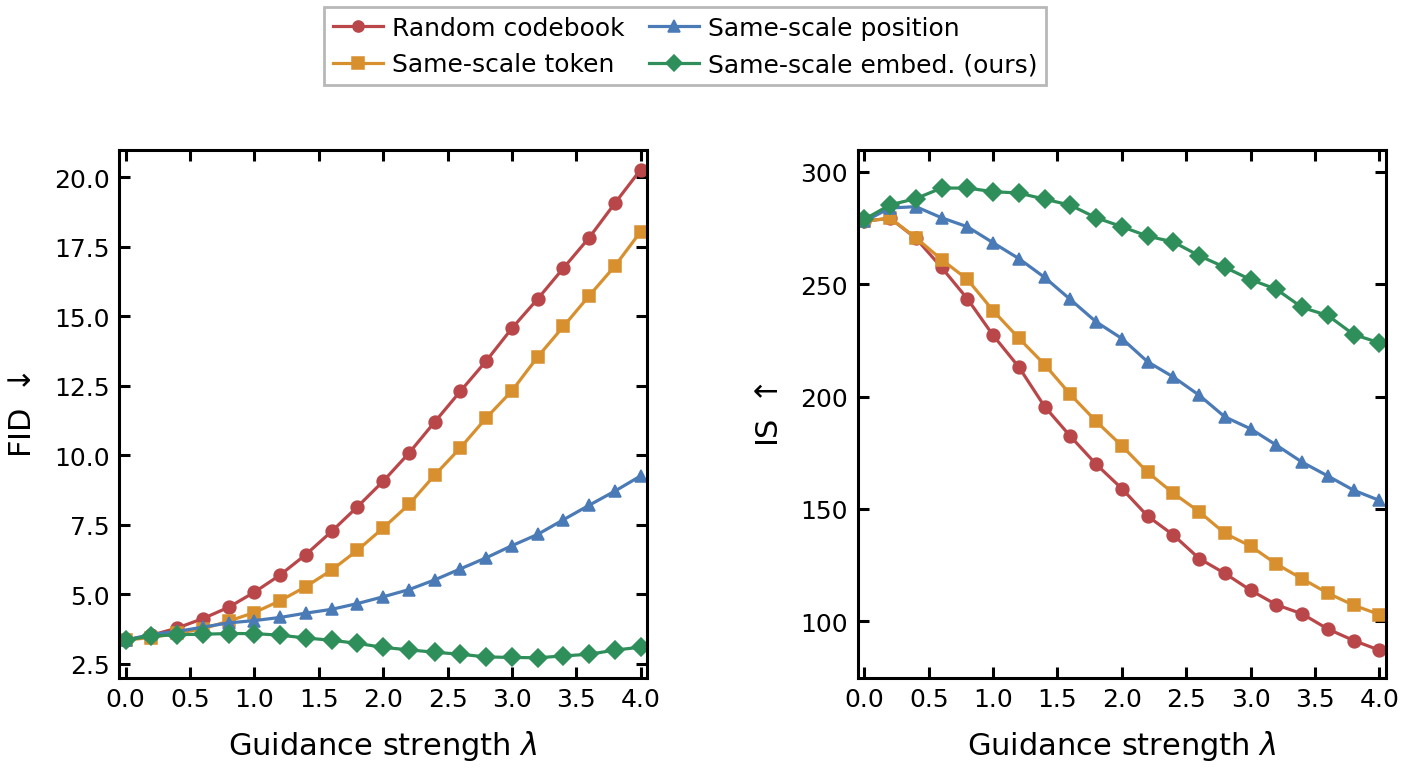}
  \caption{\textbf{FID/IS curves for corrupted-prefix replacement variants on VAR-d16.}
  Settings: VAR-d16 ImageNet sampling with $n_p{=}0.1$ and guidance-strength sweeps for random codebook, same-scale token, same-scale position, and same-scale full-embedding replacement. Same-scale full-embedding replacement is the only variant that improves FID over the unguided baseline, while incomplete or off-manifold corruptions degrade rapidly as $\lambda$ increases.}
  \label{fig:replacement-fid-is-curves}
\end{figure}

\FloatBarrier

\subsection{InfinityStar VPG schedule and latency}
\label{app:infinitystar-schedule}

Sec.~\ref{sec:experiments-infinitystar} reports a single VPG schedule for InfinityStar; we describe the schedule space and the corresponding wall-clock overhead here.

\paragraph{All-scale VPG.} Applies the generated vs.\ corrupted prefix contrast at every spacetime scale (genuine-prefix and corrupted-prefix branches), yielding $\sim$2.7$\times$ per-clip latency relative to unguided InfinityStar due to GPU memory constraints and offload / onload kv-cache of each prefix.

\paragraph{Semantic-only VPG.} Restricts the contrast to early (semantic) spacetime scales, where prompt-grounded structure is established.

\paragraph{Semantic-only scheduled VPG (Ours).} Further restricts where the corrupted-prefix branch runs across the semantic scales, lowering wall-clock overhead while preserving the semantic-axis benefit.

Among full-suite runs we tried -- unguided, all-scale VPG, and semantic-only scheduled VPG -- the semantic-only scheduled variant with $\lambda{=}0.2$ and $n_p{=}0.05$ achieves the highest leaderboard Overall ($84.35$ vs.\ $83.91$ for all-scale). All-scale prefix-contrast guidance is $\sim$2.7$\times$ slower per clip and underperforms this schedule on Overall in our runs; semantic-scale scheduling keeps overhead near $\sim$1.2$\times$ while yielding the strongest aggregate. The lower corruption probability is important for long spacetime rollouts because InfinityStar generates many more tokens than image models, so excessive prefix corruption is amplified across subsequent video predictions~\citep{InfinityStar}.

\begin{table}[!htbp]
  \caption{\textbf{VBench efficiency on InfinityStar.}
  Settings: full VBench evaluation on the InfinityStar reproduction, with overhead measured relative to unguided sampling. Semantic-only scheduled VPG reduces the all-scale overhead from $\sim$2.66$\times$ to $\sim$1.19$\times$ while giving the best Overall score among the measured schedules.}
  \label{tab:infinitystar-latency}
  \centering
  \small
  \begin{tabular}{lcccc}
    \toprule
    Method & Overall$\uparrow$ & Semantic$\uparrow$ & Latency (rel.)$\downarrow$ & Overhead$\downarrow$ \\
    \midrule
    InfinityStar (unguided) & 83.86 & 82.74 & 1.00$\times$ & 1.00$\times$ \\
    \quad + All-scale VPG & 83.91 & 82.69 & $\sim$2.66$\times$ & $\sim$2.66$\times$ \\
    \quad + Semantic-only scheduled VPG ($\lambda{=}0.2$) & 84.35 & 83.51 & $\sim$1.19$\times$ & $\sim$1.19$\times$ \\
    \bottomrule
  \end{tabular}
\end{table}

\FloatBarrier

\subsection{Data, model, and asset access}
\label{app:code-data-access}

VPG is training-free and uses no new data, model weights, or annotations. Experiments use released checkpoints and public evaluation assets from ImageNet validation statistics, GenEval, DPG-Bench, and VBench. We do not include supplementary code with the anonymous submission, but plan to open-source the VPG implementation in the near future.

\subsection{Compute resources}
\label{app:compute-resources}

All experiments were run on a single NVIDIA A800 80GB GPU.

\subsection{Existing assets, licenses, and terms}
\label{app:asset-licenses}

We use existing models, benchmarks, and evaluation code for research evaluation only, cite the original sources, and do not redistribute third-party datasets or pretrained weights. The local codebases used here include VAR, Infinity, and InfinityStar under the MIT License, and VBench under Apache-2.0. Reproduction requires obtaining datasets, checkpoints, and benchmark assets from their original providers and following their licenses and terms.

\section{Limitations}
\label{app:limitations}

VPG is an inference-time guidance rule, and its scope is therefore complementary to training-time approaches that change the learned dynamics of the generator. The corruption fraction $n_p$ and guidance scale $\lambda$ are sampling hyperparameters, analogous to the CFG scale. We find stable settings across the evaluated image and video models, but the best operating point can depend on sequence length, model family, and evaluation target. In particular, long video rollouts amplify prefix perturbations, which is why our InfinityStar setting uses a lower corruption probability and a scheduled semantic-scale application (App.~\ref{app:infinitystar-schedule}).

\section{Broader impacts}
\label{app:broader-impacts}

VPG is a sampler-level method for existing autoregressive image and video generators. Its main positive effect is practical: it can improve prompt faithfulness and visual coherence without retraining a large model, which may reduce additional training compute when a released checkpoint is already adequate for deployment or research use.

The same improvement can also increase the capability of synthetic-media systems used for deception, impersonation, spam, or disinformation. Because VPG operates at inference time and is compatible with released generators, it should inherit the safeguards of the underlying model stack, including access control, provenance or watermarking mechanisms, usage monitoring, and safety filtering. We therefore do not view VPG as changing the relevant governance requirements; rather, it strengthens the case that inference-time guidance methods should be evaluated under the same deployment and misuse standards as the generators they modify.

\section{Qualitative details and examples}
\label{app:qualitative}

This section collects qualitative material that complements the quantitative results in Sec.~\ref{sec:experiments-main}. The teaser explanation clarifies the controlled comparisons in Fig.~\ref{fig:teaser}; the VAR examples show controlled ImageNet comparisons; the InfinityStar examples show video-frame comparisons; and the DPG-Bench examples show where VPG changes Infinity generations on dense text prompts.

\subsection{Teaser figure details}
\label{app:teaser-details}

Fig.~\ref{fig:teaser} pairs controlled qualitative comparisons (left) with the conditioning-axis view that motivates VPG (right). We expand each region here.

\paragraph{Left (same-seed qualitative comparisons).} Each w/o VPG and w/ VPG pair uses the same prompt, same random seed, same frozen checkpoint, and matched sampling configuration; only the VPG branch is enabled or disabled. In the first two rows, the first two columns use the frozen Infinity~\citep{infinity} image generator. The first prompt asks for a wax seal embossed with the letters ``VPG'': without VPG, the seal reads closer to ``VDG'', while VPG recovers the intended text. The second prompt describes an owl standing among shattered mirror pieces on a floor; the baseline loses the owl under the strong broken-glass scene prior, while VPG preserves the subject. The third column uses VAR-$d30$~\citep{var} on the ImageNet~\citep{deng2009imagenet} class ``macaw'': the baseline produces an occluded, structurally weak bird, while VPG yields a coherent scarlet macaw. The last two rows use InfinityStar~\citep{InfinityStar} for text-to-video generation with the prompt: ``A towering moss-covered mech kneeling to tend a tiny rooftop bonsai garden during light rain, steam mixing with mist, tactile metal textures, contemplative mood, ultra-realistic.'' The displayed frames are sampled from same-seed clips; VPG keeps the mech, bonsai, and rooftop setting more consistently across the sequence.

\paragraph{Right (conditioning-axis motivation).} VPG and classifier-free guidance act on disjoint conditioning axes of the next-scale conditional $p_\theta(r_k\mid r_{<k},c)$. CFG exponentiates the external-condition posterior $p(c\mid r_{\le k})^{\gamma}$ to sharpen dependence on the text condition $c$, whereas VPG exponentiates the prefix posterior $p(r_{<k}\mid r_k,c)^{\lambda}$ to sharpen dependence on the generated visual prefix $r_{<k}$. Standard CFG cannot reach the prefix axis in a frozen visual AR model because the model is not trained with a null-prefix mode; VPG fills this gap with a corrupted-prefix surrogate (Sec.~\ref{sec:method-vpg}).

\subsection{VAR qualitative study}
\label{app:var-vis}

\begin{figure}[!htbp]
  \centering
  \includegraphics[width=0.86\linewidth]{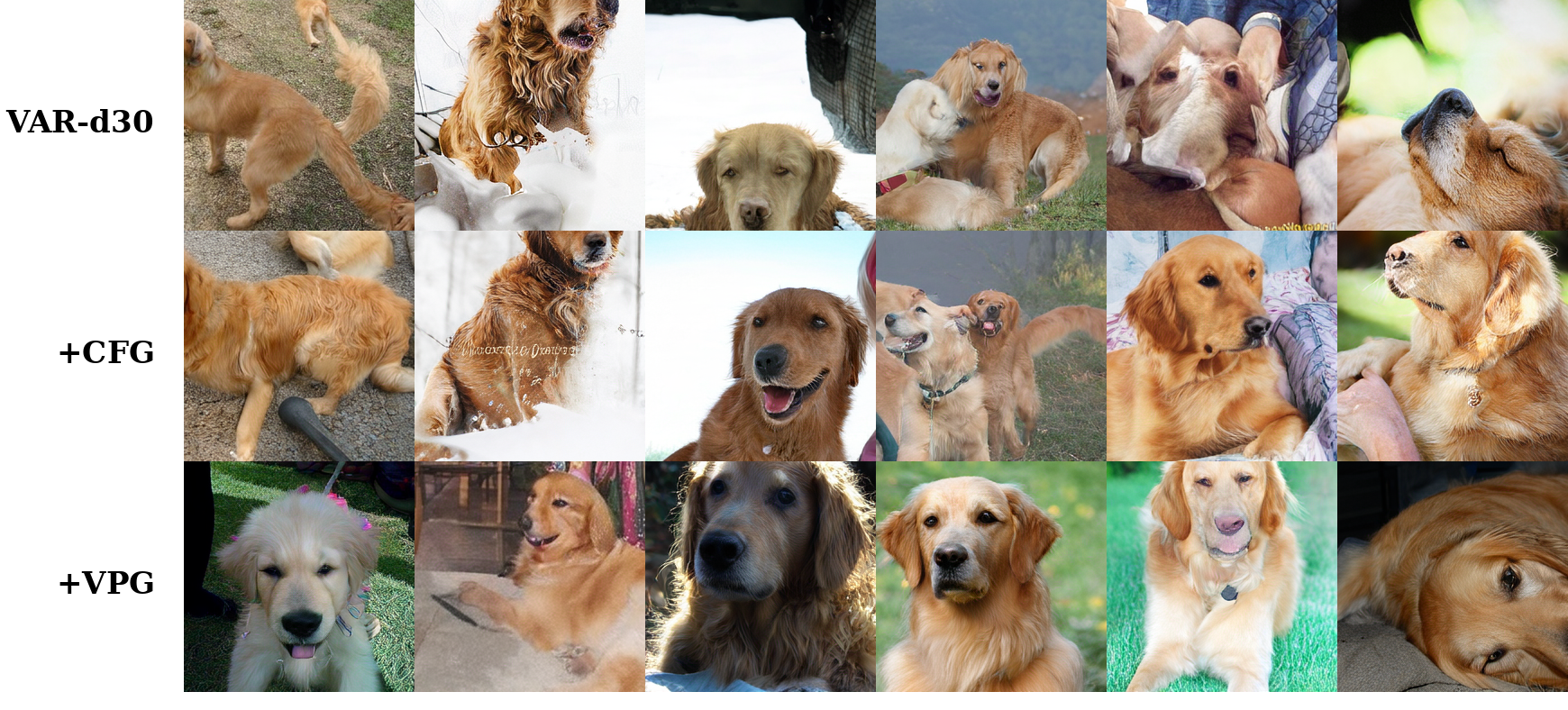}
  \caption{\textbf{VAR-$d30$ qualitative comparison for \emph{golden retriever}.}
  Settings: matched ImageNet class sampling with six samples per row; rows show VAR-$d30$, +CFG, and +VPG with $n_p{=}0.1$, $\lambda{=}1.0$. VPG improves object coherence.}
  \label{fig:var-qualitative-golden}
\end{figure}

\begin{figure}[!htbp]
  \centering
  \includegraphics[width=0.86\linewidth]{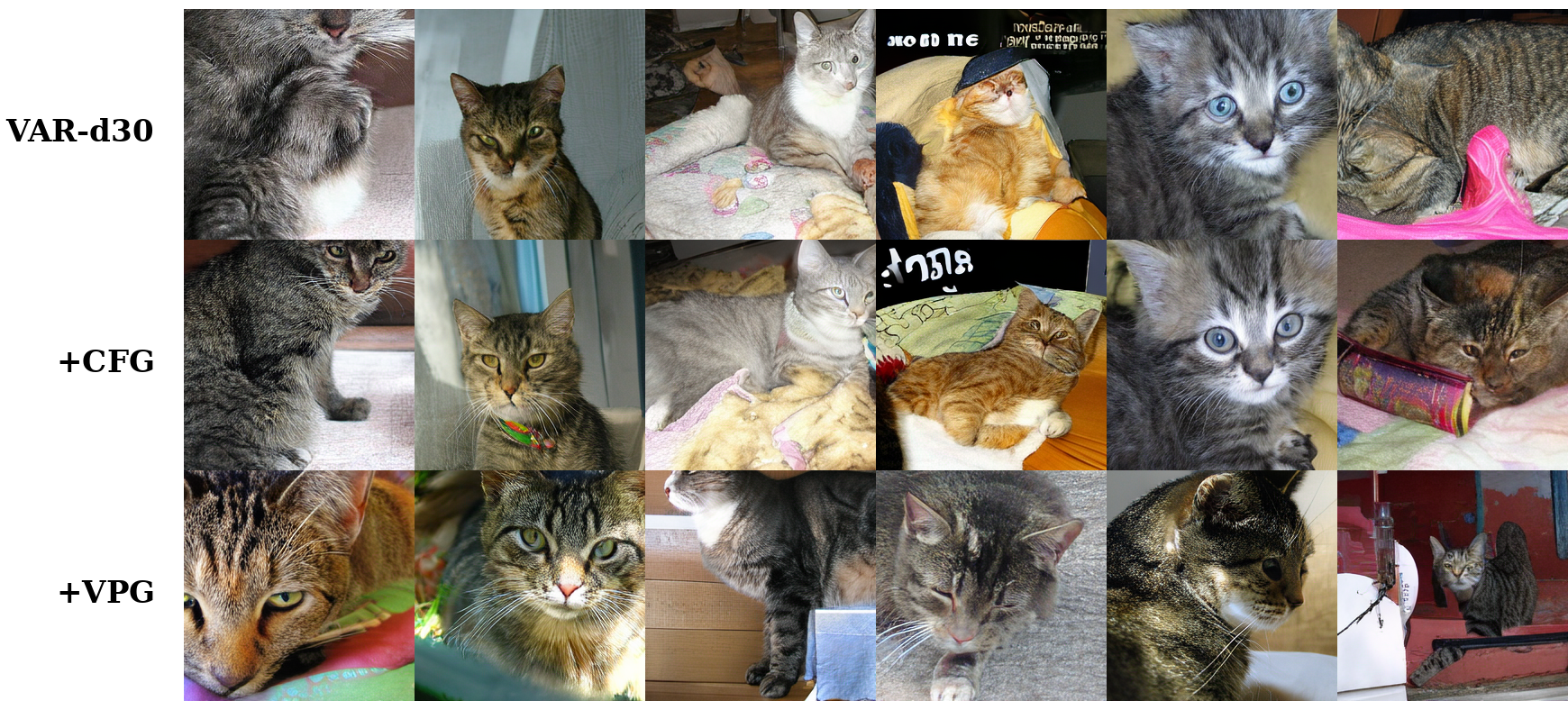}
  \caption{\textbf{VAR-$d30$ qualitative comparison for \emph{tabby cat}.}
  Settings: matched ImageNet class sampling with six samples per row; rows show VAR-$d30$, +CFG, and +VPG with $n_p{=}0.1$, $\lambda{=}1.0$. VPG yields more recognizable object structure.}
  \label{fig:var-qualitative-tabby}
\end{figure}

\begin{figure}[!htbp]
  \centering
  \includegraphics[width=0.86\linewidth]{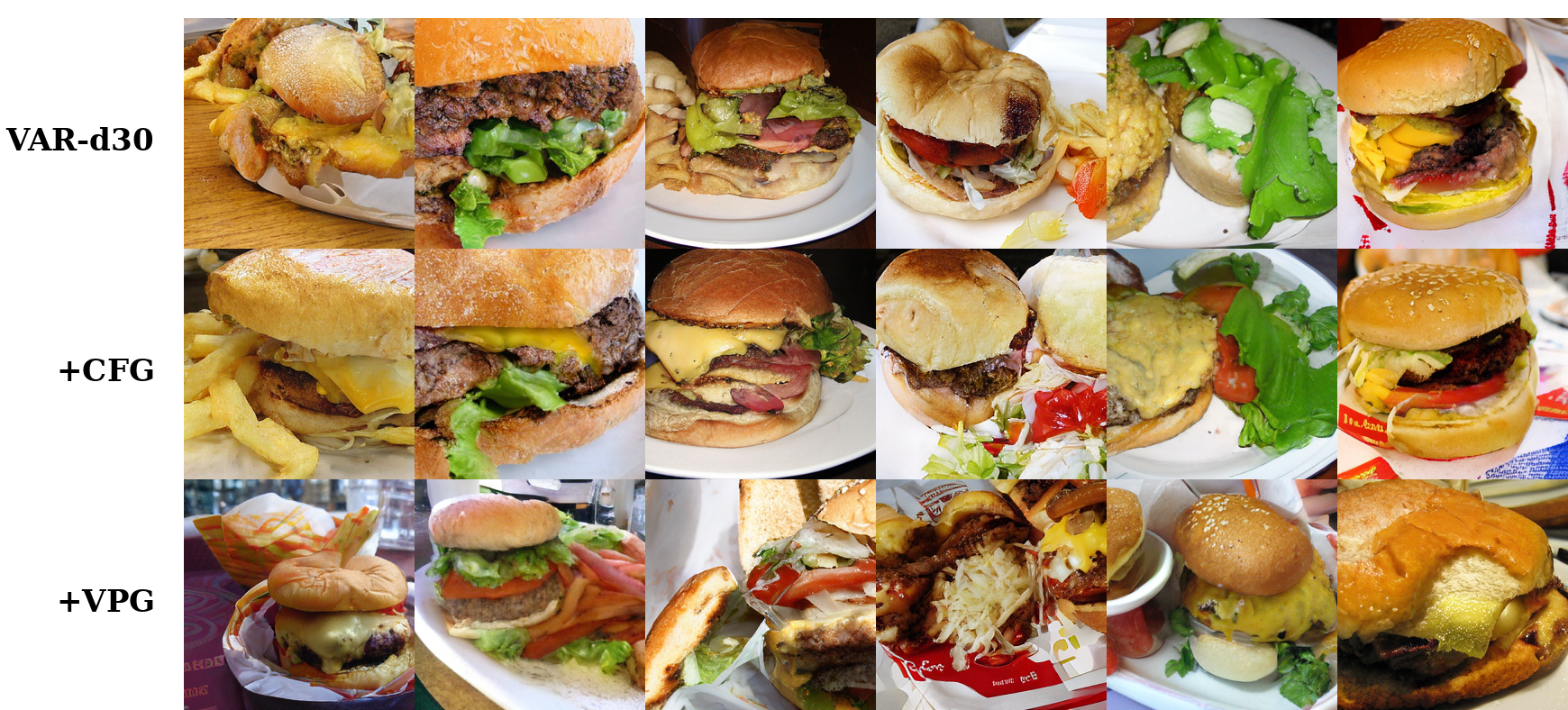}
  \caption{\textbf{VAR-$d30$ qualitative comparison for \emph{cheeseburger}.}
  Settings: matched ImageNet class sampling with six samples per row; rows show VAR-$d30$, +CFG, and +VPG with $n_p{=}0.1$, $\lambda{=}1.0$. VPG produces more stable food layouts.}
  \label{fig:var-qualitative-cheeseburger}
\end{figure}

Figs.~\ref{fig:var-qualitative-golden}--\ref{fig:var-qualitative-cheeseburger} provide same-class VAR-$d30$ qualitative comparisons on ImageNet. Each figure fixes the class and sampling configuration, with six samples per row. The rows compare the base VAR-$d30$ sampler, VAR-$d30$ with CFG, and VAR-$d30$ with both CFG and VPG.

\FloatBarrier

\subsection{InfinityStar qualitative study}
\label{app:infinitystar-vis}

Figs.~\ref{fig:infinitystar-train-library}--\ref{fig:infinitystar-paper-whale-harbor} provide matched text-to-video comparisons on InfinityStar. Each row shows five uniformly sampled frames from the generated clip; the VPG row uses semantic-only scheduled VPG with $\lambda{=}0.2$ and $n_p{=}0.05$.
{\footnotesize
Prompts:
\emph{train library}: ``A vintage train car transformed into a traveling library speeding through snowy mountains at sunrise, warm interior lamps, fluttering book pages, parallax through the windows, cinematic realism'';
\emph{subway greenhouse}: ``An abandoned underground subway platform transformed into a lush greenhouse, vines wrapping around old train cars, bioluminescent flowers pulsing as a train breeze moves the leaves, cinematic realism, rich tactile textures'';
\emph{origami whale}: ``A colossal origami whale gliding through a foggy harbor at dawn, folded paper surfaces catching golden light, tiny tugboats circling below, gulls and mist moving through the scene, poetic cinematic realism, ultra-detailed.''
\par}

\clearpage

\begin{center}
  \includegraphics[width=0.97\linewidth]{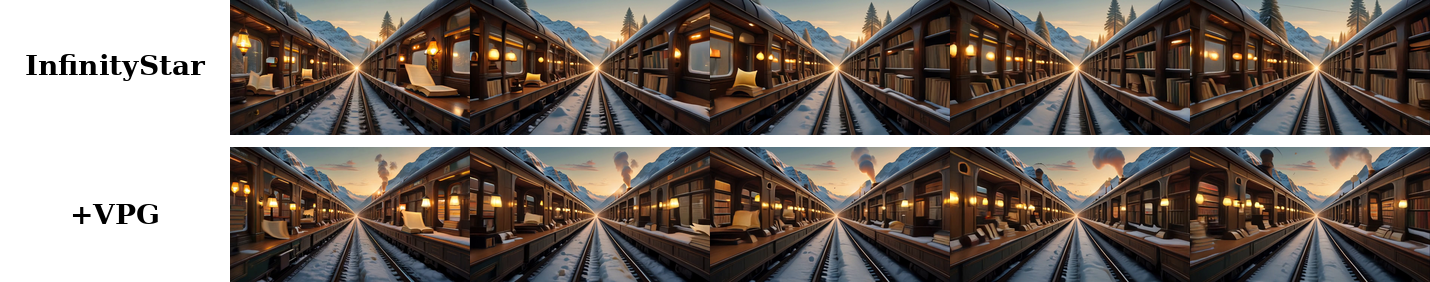}
  \refstepcounter{figure}\label{fig:infinitystar-train-library}
  {\small\textbf{Figure~\thefigure: InfinityStar qualitative comparison for the train-library prompt.}
  Settings: matched clips with five sampled frames per row; VPG uses $n_p{=}0.05$, $\lambda{=}0.2$. VPG maintains the train-library concept.}
  \vspace{0.6ex}

  \includegraphics[width=0.97\linewidth]{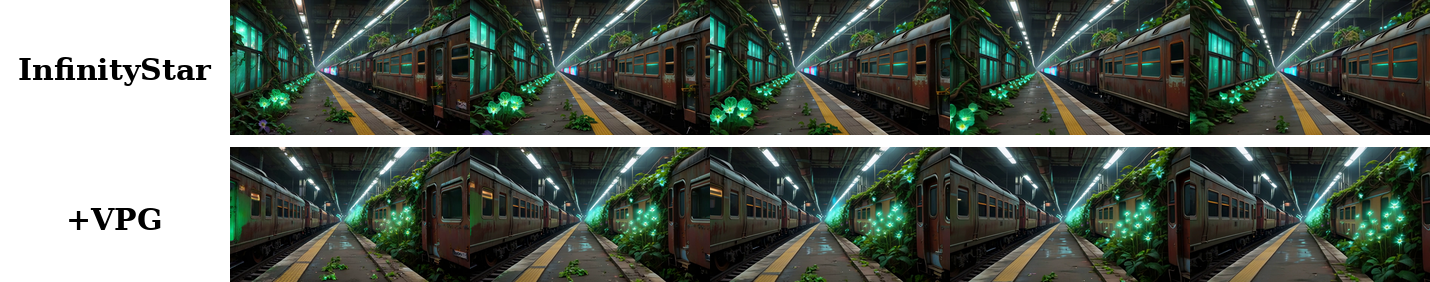}
  \refstepcounter{figure}\label{fig:infinitystar-subway-greenhouse}
  {\small\textbf{Figure~\thefigure: InfinityStar qualitative comparison for the subway-greenhouse prompt.}
  Settings: matched clips with five sampled frames per row; VPG uses $n_p{=}0.05$, $\lambda{=}0.2$. VPG better preserves vegetation and bioluminescent flower cues while maintaining the overall train-like appearance.}
  \vspace{0.6ex}

  \includegraphics[width=0.97\linewidth]{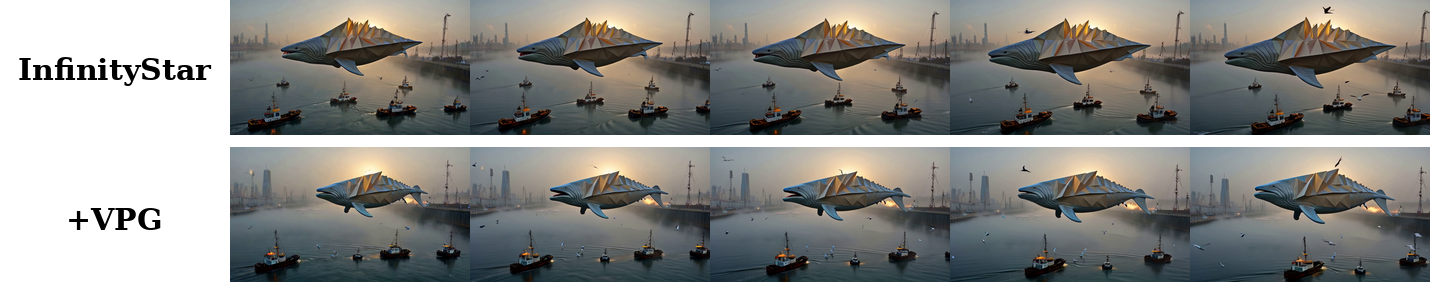}
  \refstepcounter{figure}\label{fig:infinitystar-paper-whale-harbor}
  {\small\textbf{Figure~\thefigure: InfinityStar qualitative comparison for the origami-whale prompt.}
  Settings: matched clips with five sampled frames per row; VPG uses $n_p{=}0.05$, $\lambda{=}0.2$. VPG keeps the folded-paper whale, harbor scale.}
\end{center}

\subsection{Infinity qualitative study on DPG-Bench}
\label{app:dpg-vis}

We compare the unguided Infinity~\citep{infinity} baseline with \textbf{Infinity + VPG} on dense DPG-Bench prompts. Each figure uses the same checkpoint, sampler, and seeds, with the prompt on the left, four baseline samples in the middle, and four VPG samples on the right.

\begin{figure}[!htbp]
  \centering
  \includegraphics[width=0.9\linewidth]{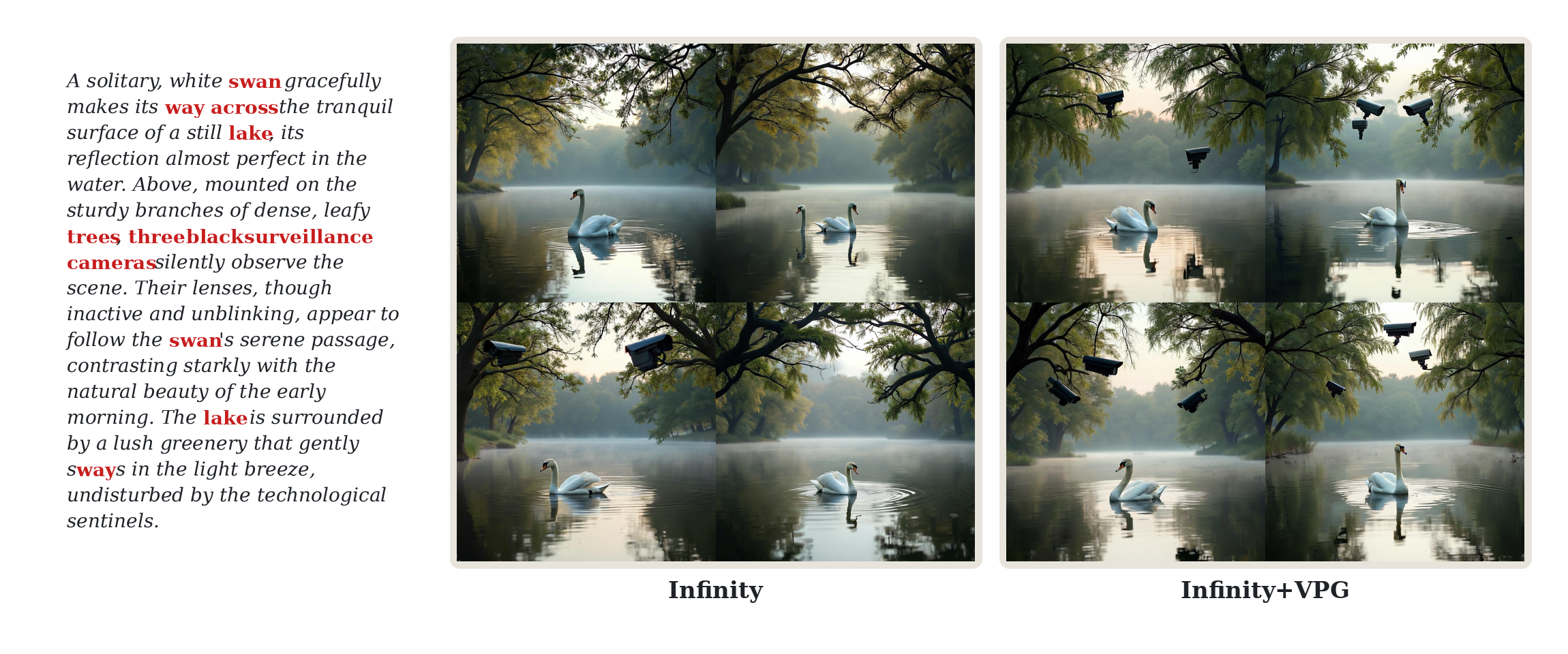}
  \caption{\textbf{DPG-Bench qualitative comparison for a surveillance-camera prompt.}
  Settings: matched Infinity baseline and Infinity + VPG generations under the same checkpoint, sampler, and seeds. VPG recovers the small cameras dropped by the baseline.}
  \label{fig:dpg-vis-01}
\end{figure}

\begin{figure}[!htbp]
  \centering
  \includegraphics[width=0.9\linewidth]{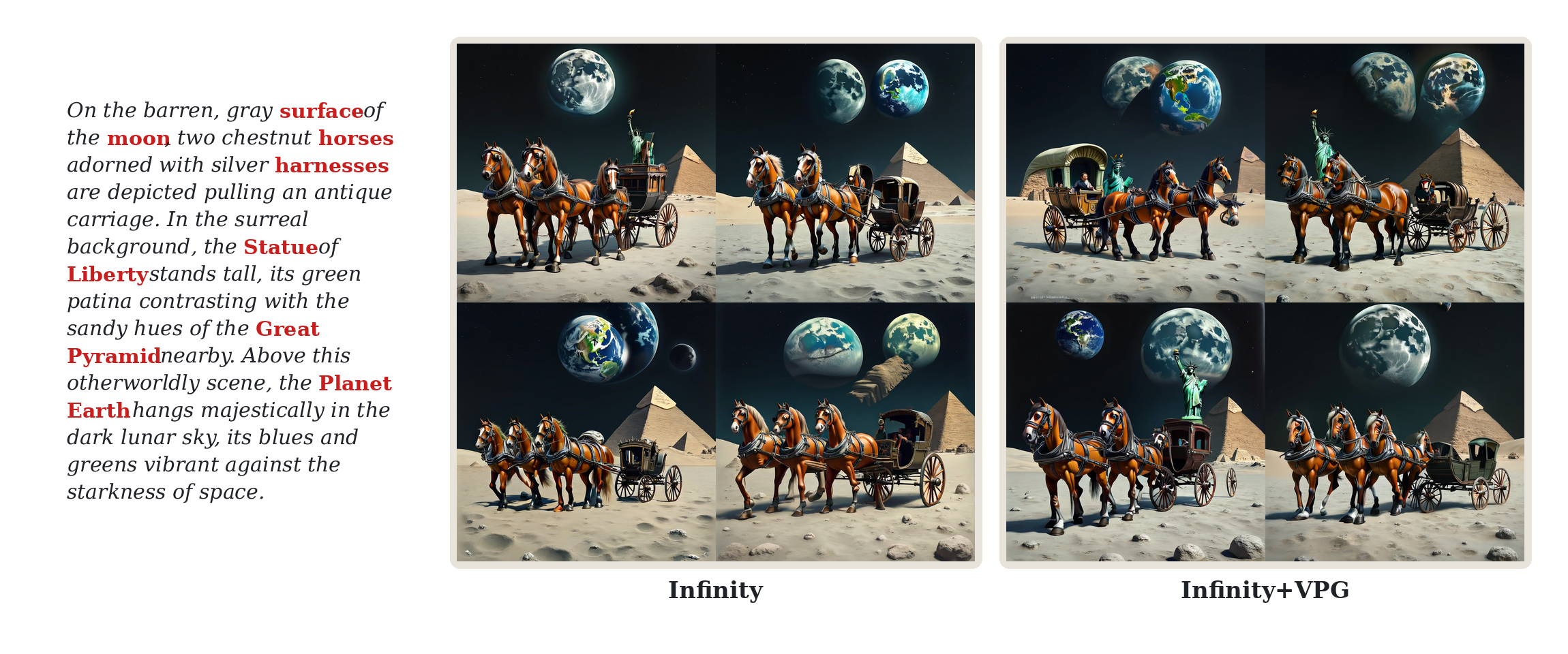}
  \caption{\textbf{DPG-Bench qualitative comparison for a lunar multi-entity prompt.}
  Settings: matched Infinity baseline and Infinity + VPG generations under the same checkpoint, sampler, and seeds. VPG better preserves distinct entities and relative positions.}
  \label{fig:dpg-vis-07}
\end{figure}

\begin{figure}[!htbp]
  \centering
  \includegraphics[width=0.9\linewidth]{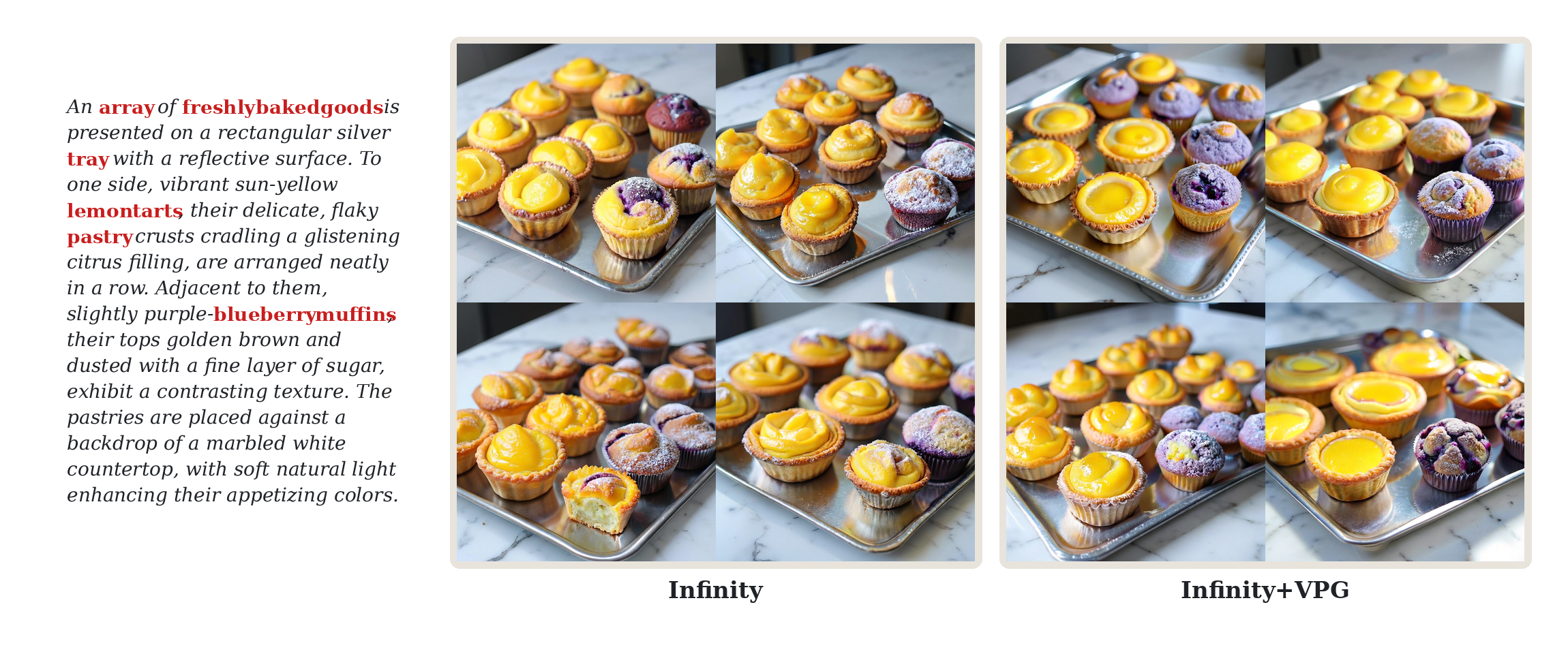}
  \caption{\textbf{DPG-Bench qualitative comparison for a two-food prompt.}
  Settings: matched Infinity baseline and Infinity + VPG generations under the same checkpoint, sampler, and seeds. VPG renders both food groups side by side.}
  \label{fig:dpg-vis-14}
\end{figure}

\FloatBarrier

\begin{figure}[!t]
  \centering
  \includegraphics[width=0.9\linewidth]{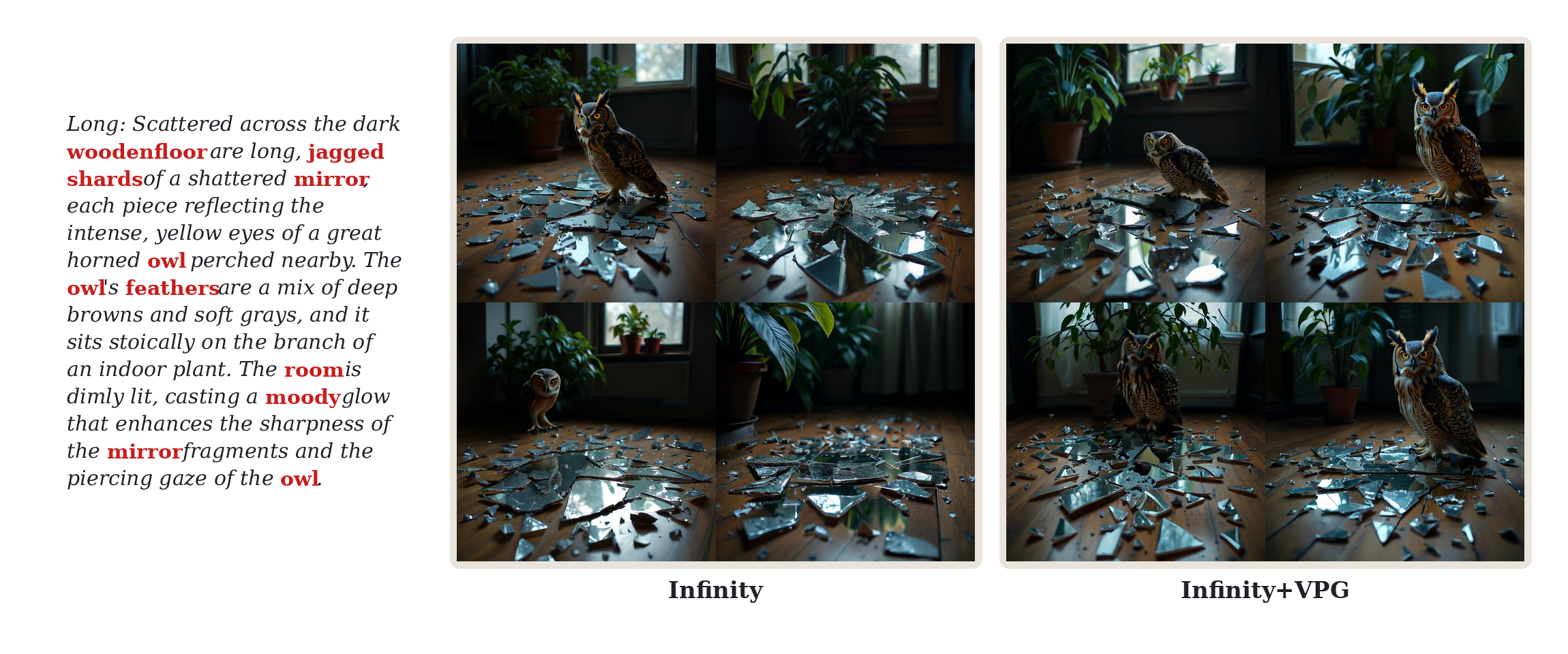}
  \caption{\textbf{DPG-Bench qualitative comparison for an owl-and-mirror prompt.}
  Settings: matched Infinity baseline and Infinity + VPG generations under the same checkpoint, sampler, and seeds. VPG produces clearer mirror fragments.}
  \label{fig:dpg-vis-19}
\end{figure}

Across all four prompts, the qualitative trend matches Tab.~\ref{tab:infinity-geneval-dpg}: VPG helps when the unguided model omits or merges minor entities under a strong scene prior.

\clearpage

\end{document}